\documentclass[acmsmall]{acmart}

\AtBeginDocument{%
  \providecommand\BibTeX{{%
    \normalfont B\kern-0.5em{\scshape i\kern-0.25em b}\kern-0.8em\TeX}}}

\setcopyright{acmcopyright}
\copyrightyear{2019}
\acmYear{2019}
\acmDOI{10.1145/1122445.1122456}

\newcommand{\hide}[1]{}

\newcommand{\etal}{\textit{et al.}~}
\usepackage{xcolor}
\usepackage{float}

\usepackage{multirow}

\usepackage{tcolorbox}

\newtcolorbox{mybox}[3][]
{
	colframe = #2!25,
	colback  = #2!10,
	coltitle = #2!20!black,  
	title    = #3,
	#1,
}

\graphicspath{{./img/}}

\usepackage{amsmath}

\newcommand{\punctualcolor}{3166FF}
\newcommand{\teliccolor}{009901}
\newcommand{\ateliccolor}{F8A102}

\begin{document}

\title{Am I done? Predicting action progress in videos}

\author{Federico Becattini}
\affiliation{%
  \institution{University of Florence}
  \city{Florence}
  \country{Italy}}
\email{federico.becattini@unifi.it}

\author{Tiberio Uricchio}
\affiliation{%
  \institution{University of Florence}
  \city{Florence}
  \country{Italy}}
\email{tiberio.uricchio@unifi.it}

\author{Lorenzo Seidenari}
\affiliation{%
  \institution{University of Florence}
  \city{Florence}
  \country{Italy}}
\email{lorenzo.seidenari@unifi.it}

\author{Lamberto Ballan$^\ast$}
\affiliation{%
  \authornote{This work was partially done while this author was at the University of Florence, Italy.}
  \institution{University of Padova}
  \city{Padova}
  \country{Italy}}
\email{lamberto.ballan@unipd.it}

\author{Alberto Del Bimbo}
\affiliation{%
  \institution{University of Florence}
  \city{Florence}
  \country{Italy}}
\email{alberto.delbimbo@unifi.it}

%

\renewcommand{\shortauthors}{F.~Becattini et al.}

\begin{abstract}
  In this paper we deal with the problem of predicting action progress in videos. We argue that this is an extremely important task since it can be valuable for a wide range of interaction applications.
  To this end we introduce a novel approach, named ProgressNet, capable of predicting \textit{when} an action takes place in a video, \textit{where} it is located within the frames, and \textit{how far} it has progressed during its execution. 
  To provide a general definition of action progress, we ground our work in the linguistics literature, borrowing terms and concepts to understand which actions can be the subject of progress estimation. As a result, we define a categorization of actions and their phases.
  Motivated by the recent success obtained from the interaction of Convolutional and Recurrent Neural Networks, our model is based on a combination of the Faster R-CNN framework, to make frame-wise predictions, and LSTM networks, to estimate action progress through time.
  After introducing two evaluation protocols for the task at hand, we demonstrate the capability of our model to effectively predict action progress on the UCF-101 and J-HMDB datasets.
\end{abstract}



\maketitle

\section{Introduction}
\label{sec:intro}

\begin{figure}[t]
	\centering
	\begin{tcolorbox}[colframe=green, left=0pt, right=0pt,top=0pt, bottom=-12pt]
		\includegraphics[width=.99\columnwidth]{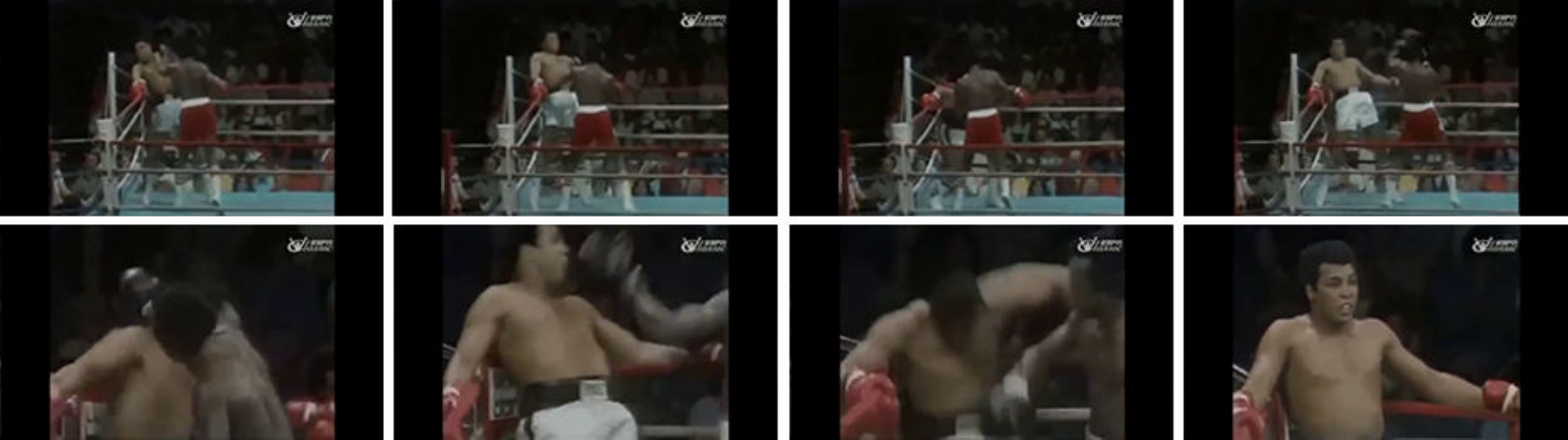} \\
	\end{tcolorbox}
	\vspace{-3px}
	\begin{tcolorbox}[colframe=red, left=0pt, right=0pt,top=0pt, bottom=0pt]
		\includegraphics[width=1\columnwidth]{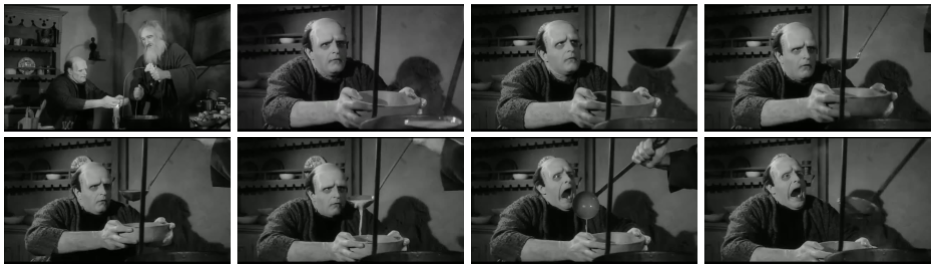} 
	\end{tcolorbox}
	\caption{The ability to estimate action progress is essential for interaction. In 1977 progress estimation allowed the boxer Muhammad Ali to dodge 21 punches in 10 seconds: \protect\url{https://youtu.be/nxZ-J7xit5Y}. In the movie ``Young Frankenstein'' instead, the monster is not able to estimate progress and react accordingly: \protect\url{https://goo.gl/muYbCb}}
	\label{pullfig}
\end{figure}

Humans are not only able to recognize actions and activities, but they can also understand how far an action has progressed and make important decisions based on this information.
From simple choices, like crossing the street when cars have passed, to more complex activities like intercepting the ball in a basketball game, an intelligent agent has to recognize and understand how far an action has advanced at an early stage, based only on what it has seen so far.
If an agent has to act to assist humans, it can not wait for the end of the action to perform the visual processing and act accordingly (Fig.~\ref{pullfig} shows an example sequence for this phenomena). 
Therefore, the ultimate goal of action understanding should be the development of an agent equipped with a fully functional perception action loop, from predicting an action before it happens, to following its progress until it ends.
This is supported also by experiments in psychology showing that humans continuously understand the actions of others in order to plan their goals \cite{flanagan2003action}. 
Consequently, a model that is able to forecast action progress would enable new applications in robotics (e.g. human-robot interaction, realtime goal definition) and autonomous driving (e.g. avoid road accidents). 

Broadly speaking, our work falls into the area of \emph{predictive vision}, an emerging field which is gaining much interest in the recent years. Several approaches have been proposed to perform prediction of the near future, be that of a learned representation \cite{vondrick2016fast,lee2017sort}, a video frame \cite{vondrick2016generating,mathieu2015deep}, or directly the action that is going to happen \cite{fermuller2016prediction,liang2019peeking}.
However, we believe that fully solving action understanding requires not only to predict the future outcome of an action, but also to understand what has been observed so far in the progress of an action. 
As a result, in this paper we introduce the novel task of predicting \emph{action progress}, \emph{i.e.}~the prediction of how far an action has advanced during its execution. In other words, considering a partial observation of some human action, in addition to understanding \textit{what} action and \textit{where} it is happening, we want to infer \textit{how long} this action has been executed for with respect to its duration.
As a simple example of application, let us consider the use case of a social robot trained to interact with humans. The correct behaviour to respond to a handshake would be to anticipate, with the right timing, the arm motion, so as to avoid a socially awkward moment in which the person is left hanging. This kind of task cannot be solved unless the progress of the action is estimated accurately.

Some closely related problems have been recently addressed by the computer vision community.
First of all, predicting action progress is conceptually different from action recognition and detection \cite{ggjm2015tubes,serena2016fgv,zhao2017temporal,xu2019temporal}, where the focus is on finding \textit{where} the action occurred in time and space.
Action completion \cite{damen2016rgbc,xiong2017pursuit,yuan2017temporal,zhao2017temporal,Heidarivincheh2018ActionCA,heidarivincheh2019weakly} is indeed a related task, where the goal is to predict when an action can be classified as complete to improve temporal boundaries and the classifier accuracy on incomplete sequences. However, this is easier than predicting action progress because it does not require to predict the partial progress of an action.
Action progress prediction is an extremely challenging task since, to be of maximum utility, the prediction should be made online while observing the video. While a thick crop of literature addresses action detection and spatio-temporal localization \cite{actoms,weinzaepfel2015learning,serena2016fgv,hou2017tube,gao2017turn,zhu2017tornado,peng2016multi,Kalogeiton2017localization,long2019gaussian}, predicting action progress is more closely related to online action detection \cite{hoai2014ed,uva2016online,singh2017online,kong2017deep,xu2019temporal}.
Here the goal is to accurately detect, as soon as possible, when an action has started and when it has finished, but they do not have a model to estimate the progress.
Some similarities are shared with the task of activity recognition \cite{twinanda2018rsdnet,LiProgressSurgery} where the goal is to detect which high-level phase is currently in progress in a long procedure composed by multiple actions. Differently, action progress focus on such actions which can be made of one or many movements but can be detected by visual means only. That is harder than activity recognition where the phases represent states of a process and can also be inferred by using a knowledge base or by the objects that are in use. 

Not every action has a progress to be estimated. For instance, there are actions which are instantaneous (such as hitting a ball) or which do not have a clear goal given the available information (such as walking without a precisely defined destination). Thus, a classification of addressable actions is needed to successfully model action progress. Unfortunately, such a formal classification is a still an open research subject, but there is a building evidence in neuroscience that language is strongly linked to actions and is correlated to similar grammars \cite{steele2012action,arbib2006sentence,pastra2012minimalist}. 
As a result, in this work we take inspiration from the linguistics literature which extensively discuss the topic of actions and the category of verbs used to refer to them. We propose a unified view of the problem discussing how progress can be estimated according to the corresponding verb classification.


In summary, in this paper we propose ProgressNet, a model for action progress prediction of multiple actors using a supervised recurrent neural network fed with convolutional features. The main contributions of this work are the following:
\begin{itemize}
	\item We define the new task of action progress prediction, which we believe is a fundamental problem in developing intelligent planning agents. We take inspiration from linguistics literature to classify which actions are suitable for the task. We also design and present an experimental protocol to assess performance.
	\item Given that actions are often composed of sub action phases, we propose two formal models of action progress. A simple model which considers actions in their entirety and one which introduces a sequence of sub phases. 
	\item We present a holistic approach capable of predicting action progress while performing spatio-temporal action detection. ProgressNet can be naturally fitted into any online action detector model. To encourage precise prediction of progress, we also contribute a novel Boundary Observant loss which penalizes prediction errors on temporal boundaries. 
\end{itemize}



\section{Related Work}\label{sec:related}

Human action understanding has been traditionally framed as a classification task~\cite{poppe2010,csur2011}.
However, in recent years, several works have emerged aiming at a more precise semantic annotation of videos, namely action localization, completion and prediction.

Frame level action localization has been tackled extending state-of-the art object detection approaches \cite{ren2015faster} to the spatio-temporal domain. A common strategy is to start from object proposals and then perform object detection over RGB and optical flow features using convolutional neural networks \cite{ggjm2015tubes,peng2016multi,saha2016deep}.
Gkioxari \etal generate action proposals by filtering Selective Search boxes with motion saliency, and fuse motion and temporal decision using an SVM \cite{ggjm2015tubes}.
More recent approaches, devised end-to-end tunable architectures integrating region proposal networks in their model \cite{peng2016multi,saha2016deep,shou2017cvpr,sst_buch_cvpr17,Kalogeiton2017localization,hou2017tube,singh2017online,zhao2017temporal,gao2017turn,long2019gaussian}. As discussed in \cite{saha2016deep}, most action detection works do not deal with untrimmed sequences and do not generate action tubes. To overcome this limitation, Saha \etal \cite{saha2016deep} propose an energy maximization algorithm to link detections obtained with their framewise detection pipeline. Another way of exploiting the temporal constraint is to address action detection in videos as a tracking problem, learning action trackers from data \cite{weinzaepfel2015learning}. 
To allow online action detection, Singh \etal \cite{singh2017online} adapted the Single Shot Multibox Detector \cite{liu2016ssd} to regress and classify action detection boxes in each frame. Then, tubes are constructed in real time via an incremental greedy matching algorithm. Considering that actions may have different temporal scales, in \cite{long2019gaussian}, the authors go beyond predetermined temporal scales and propose to use Gaussian kernels to dynamically optimize them. 

Approaches concentrating in providing starting and ending timestamps of actions have been proposed \cite{actoms,serena2016fgv,jcn2016fast,shou2017cvpr,escorcia2020guess,liu2019completeness,nguyen2019weakly}. Heilbron \etal\cite{jcn2016fast} have recently proposed a very fast approach to generate temporal action proposals based on sparse dictionary learning.
Yeung \etal \cite{serena2016fgv} looked at the problem of temporal action detection as joint action prediction and iterative boundary refinement by training a RNN agent with reinforcement learning.
In Shou \etal \cite{shou2017cvpr} a 3D convolutional neural network is stacked with Convolutional-De-Convolutional filters in order to abstract semantics and predict actions at the frame-level granularity. They report an improved performance in action detection frame-by-frame, allowing a more precise localization of temporal boundaries.
More recently, this line of research evolved into the stricter problem of locating an action having only a video-level label available \cite{zhao2017temporal,liu2019completeness,nguyen2019weakly,escorcia2020guess}. Zhao \etal \cite{zhao2017temporal} introduced an explicit modelization of starting, intermediate and ending phases via structured temporal pyramid pooling for action localization. They show that this assumption helps to infer the completeness of the proposals. In \cite{liu2019completeness,nguyen2019weakly}, the action-context separation is explicitly considered. The idea is that parts of the video where an action has not taken place can be used as context of the instance, modeled with an end-to-end  \cite{liu2019completeness} or probabilistic \cite{nguyen2019weakly} approach, to better separate the temporal edges. Escorcia \etal \cite{escorcia2020guess} specialize on localizing human actions by considering actor proposals derived from a detector for human and non-human actors intended for images. They use an actor-based attention mechanism, which is end-to-end trainable. A distinct line of research considers explicitly the temporal dimension \cite{hou2017tube,Kalogeiton2017localization} either using 3D ConvNets on tubes \cite{hou2017tube} or exploiting multiple frames to generate a tube \cite{Kalogeiton2017localization}. 
However, all these methods do not understand and predict the progress of actions, but only the starting and ending points. Differently from them, we explicitly model and predict action progress during the entire action and not only the starting and ending boundaries.

Orthogonal approaches to ours, that implicitly model action progress cues can be found in action anticipation \cite{shugao2016fast,aliakbarian2017encouraging,uva2016online,hoai2014ed} and completion \cite{damen2016rgbc,xiong2017pursuit,yuan2017temporal,zhao2017temporal,soran2015gnma,Heidarivincheh2018ActionCA,heidarivincheh2019weakly}. 
The former aims at predicting actions before they start or as soon as possible after their beginning, the latter instead binarizes the problem by predicting if an ongoing action is finished or not. Early event detection was first proposed by Hoai \etal \cite{hoai2014ed}. In \cite{hoai2014ed}, a method based on Structured Output SVM is proposed to perform early detection of video events. To this end, they introduce a score function of class confidences that has a higher value on partially observed actions. In \cite{kong2017deep}, the same task is tackled using a very fast deep network. Interestingly, the authors noted that the predictability of an action can be very different, from instantly to lately predictable. In \cite{shugao2016fast} an LSTM is used to obtain a temporally increasing confidence to discriminate between similar actions in early stages while Aliakbarian \etal \cite{aliakbarian2017encouraging} combine action and context aware features to make predictions when a very small percentage of the action is observed.
In \cite{damen2016rgbc} RGB-D data is used to discriminate between complete and incomplete actions and \cite{soran2015gnma} tries to detect missing sub activities to timely remind them to the user.
More recently, Heidarivincheh \etal \cite{Heidarivincheh2018ActionCA} presented an LSTM-based architecture which is able to predict the frame's relative position of the completion moment by either classification or regression. This problem was also further addressed by the same authors in a weakly supervised setting \cite{heidarivincheh2019weakly}.

The growing area of predictive vision \cite{mathieu2015deep,vondrick2016fast,wang2016transf,vondrick2016generating} is also related to action progress prediction. Given an observed video, the goal is to obtain some kind of prediction of its near future. 
Vondrick \etal \cite{vondrick2016fast} predict a learned representation and a semantic interpretation, while subsequent works predict the entire video frame \cite{vondrick2016generating,mathieu2015deep}. 
All these tasks are complementary to predicting action progress since, instead of analyzing the progress of an action, they focus on predicting the aftermath of an action based on some preliminary observations.
A few recent works have addressed tasks that are very close to action progress prediction.
Neumann and Zisserman \cite{neumann2019} framed the problem of future event prediction by defining a number of representations and loss functions to detect if and when a specific event will occur. To this end, they attempt to predict and regress the time to event probability which is highly related to predicting action progress.
In \cite{han17grammars} a residual action recognition model is used to estimate the progress of human activities to implicitly learns a temporal task grammar with respect to which activities can be localized and predicted. 

Finally, a few preliminary attempts to estimate activity progress have been presented in the medical domain \cite{patra18,twinanda2018rsdnet} aiming at predicting remaining surgery durations or anatomical motions. Nonetheless, these approaches are limited to predicting which action is in progress while doing a durative activity made by a sequence of actions, without explicitly predicting how much each action is completed.

\begin{figure*}[t!]
	\centering
	\includegraphics[width=.995\textwidth]{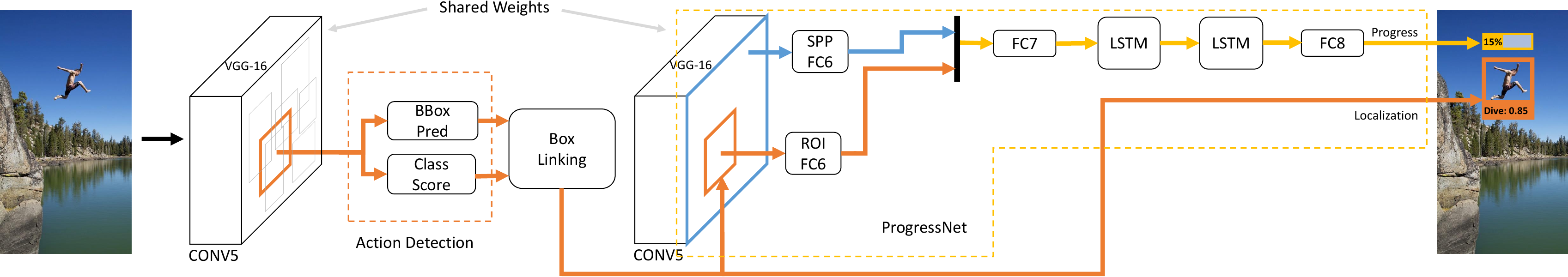}
	\caption{Proposed Architecture. On the left (highlighted in orange), we show the classification and localization data flows for tube generation. On the right (highlighted in yellow), our ProgressNet. Region (ROI FC6) and Contextual features (SPP FC6) from the last convolutional map are concatenated and then fed to a Fully Connected layer (FC7). Two cascaded LSTMs perform action progress prediction.}\label{fig:arch}
\end{figure*}



\section{Action Progress}\label{sec:approach}

In addition to categorizing actions (\emph{action classification}), identifying their boundaries spanning through the video (\emph{action detection}) and localizing the area where they are taking place within the frames (\emph{action localization}), our goal is to learn to predict the progress of an ongoing action.
We refer to this task as \emph{action progress prediction}.


Providing a definition of action progress is not trivial. In the following we analyze the problem thoroughly and propose two definitions to model progression: \textit{(i)} a linear interpretation which is versatile and can be applied to any sequence annotated for action detection; \textit{(ii)} a phase-based interpretation where actions are precisely split into sub-events and manually annotated to obtain a richer representation that captures non-linear dynamics.

In order to obtain a clear and well reasoned taxonomy of actions and a comprehensive representation of action progress, we borrow a few concepts from verb and action classifications in linguistics. These classifications are helpful to understand for which actions progress may be defined and thus to derive appropriate models. According to \cite{vendler1957verbs}, verbs can be classified into four categories: those that express \textit{activity}, \textit{accomplishment}, \textit{achievement} and \textit{state}. For our goal, we are not interested in stative verbs, which are used to describe the truth of a certain property (e.g. "someone knows something" describes the state of knowledge and "to know" is considered a stative verb) and we are instead interested in the remaining categories since they all describe actions.
About actions, in \cite{comrie1976aspect} a distinction is made between \textit{punctual} and \textit{durative} actions and whether actions have a defined goal or not (referred to as \textit{telic} and \textit{atelic}, respectively) (see Tab. \ref{tab:comrie}). A logic test to identify telic from atelic actions is the following: \textit{SUBJECT was VERB-ing until something happened and he/she stopped VERB-ing. Did he/she VERB?}

Example: 
\begin{itemize}
	\item Mark was \textit{walking} until Bill made him fall. Did Mark \textit{walk}? Yes $\rightarrow$ \textbf{Atelic} 
	\item Susan was \textit{shooting} a basketball until Anne made her fall. Did Susan \textit{shoot} the ball? No $\rightarrow$ \textbf{Telic}
\end{itemize}

Progress can be defined clearly when the action implies a change of state on the actor. 
Therefore, durative telic actions are the most appropriate for our task, since they have a duration in time and a goal that clearly implies a change of state. Punctual actions, being telic or atelic, do not progress since they do not have a clear time extent i.e. they happen instantaneously. 

Durative actions can possibly be decomposed in phases. Phases might have a different classification from the whole action. There are certain activities that are atelic if looked as a whole but can be decomposed in telic phases. Walking for instance can be reformulated as a sequence of movements of legs and feet: rise right foot, move leg, land right foot, rise left foot and so on. All these sub actions are telic-durative while the whole action of walking is atelic-durative. Similarly, telic actions might contain phases with an atelic nature. For instance, the execution of the telic action Pole Vault always has the same structure, made of telic and atelic phases: running (\textit{atelic - durative}), sticking the pole in the ground (\textit{telic - punctual}), jumping over the bar (\textit{telic - durative}), touching ground (\textit{telic - punctual}), standing up (\textit{telic - durative}). 

For the scope of our research there are two main problems that make it hard to develop a model of action progress. The first is the intrinsic difficulty of annotating temporal boundaries, for any type of action \cite{sigurdsson2017}. The second is the difficulty of splitting durative actions into phases, collecting prior information regarding their structure or subparts. From a computer vision point of view is hard to correctly annotate every single step of a walking action and it can be considered even harder to train a classifier to effectively recognize all single short spanned events of such kind.

Based on these considerations we have developed two distinct models of progression that are suited for distinct cases: (i) a linear interpretation that applies to actions regarded as a whole; (ii) a phase-based interpretation where actions are precisely split into sub-events and manually annotated to obtain a richer representation that captures non-linear dynamics. Both interpretations are built upon the concept of \textit{action tube}~\cite{ggjm2015tubes,saha2016deep}, i.e. a sequence of bounding boxes spanning from a starting frame $f_S$ to an ending frame $f_E$ and enclosing the subject performing the action. It is important to tie the definition of progress with a specific action tube, rather than a full sequence, since there might be more than an agent performing an action in the same frame with different progress ratios. We predict the progress values online, frame-by-frame, by only observing the past frames of the tubes from $t_S$ to the current frame $t_i$.

\begin{table}[t]
	\footnotesize
	\begin{tabular}{c|cc}
		& Punctual & Durative \\ \hline
		Telic  & Achievement (\textit{to release}) & Accomplishment (\textit{to drown}) \\ \hline
		Atelic & Semelfactive (\textit{to knock})  & Activity (\textit{to walk})  \\ \hline
		Changeless & - & State (\textit{to know})  \\ 
	\end{tabular}
	\caption{Comrie's categories of verbs. \cite{comrie1976aspect} }
	\label{tab:comrie}
\end{table}

\subsection{Linear progress}\label{subsec:linearprog}

We provide a first definition of action progress with a linear interpretation.
Given a frame $f_i$ belonging to an action tube in [$t_S,t_E$], action progress can be interpreted as the fraction of the action that has already passed.
Therefore, for each box in a tube at time $t_i$, we define the target action progress as:
\begin{equation}
p_i  = \frac{t_i - t_S}{t_E - t_S} \in [0,1],
\label{eq:progress}
\end{equation}

This definition is versatile and can be applied to any sequence annotated for action detection. A similar linear modeling of action evolution has been previously used in action anticipation \cite{aliakbarian2017encouraging} and in action synchronization \cite{dwibedi2019temporal} with encouraging results.
Although simple, a key advantage of this definition is that it does not require to define any prior information regarding the structure or subparts of an action, making it applicable to a large set of actions. As a result, we can learn predictive models from any dataset containing spatio-temporal boundary annotations. This means that the task of action progress in its linear interpretation does not require to collect additional annotations for existing datasets, since the action progress values can be directly inferred from the temporal annotations. This is a major strength, also considering the intrinsic difficulty of annotating action temporal boundaries \cite{sigurdsson2017}. 
While this model is best suited for durative telic actions, in the experimental section we show that it is also reasonable in many cases, including durative atelic actions.

\begin{figure*}[t]
	\centering
	\begin{tabular}{cc}
		\includegraphics[width=.47\textwidth]{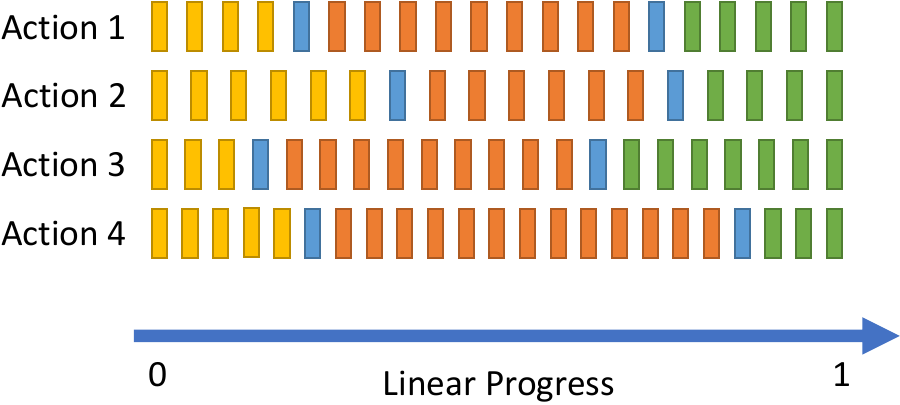} & \includegraphics[width=.47\textwidth]{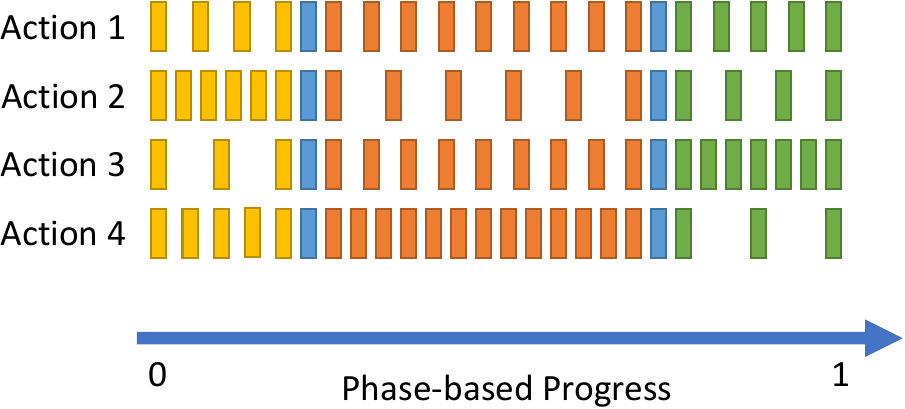}
	\end{tabular}
	\caption{Linear progress vs. Phase-based progress. Each rectangle represents a frame in an action tube. Different colors correspond to different phases. Blue rectangles represent punctual phases, which act as boundaries for durative phases. In the phase-based definition, progress is defined in order to align phase boundaries across different executions of the same action.}
	\label{fig:progresses}
\end{figure*}

\subsection{Phase-based progress}
\label{sec:phasebased}
Despite the simplicity and effectiveness of adopting a linear interpretation of progress, certain actions may indeed exhibit dilations or contractions in the execution rate, yielding to a nonlinear progression. This behavior is observable when the action can be broken down into a sequence of sub-phases that defines its structure.

Additionally, in certain cases the structure is clear and sequential while in other cases it is not. Usually telic actions exhibit this structure. On the other hand, atelic actions, such as \textit{Dancing}, may be defined by a random sequence of moves with no identifiable structure, which can even follow cyclic or erratic patterns.

The phase-based progress formulation provides a better definition of action progress. In this approach, we first split a durative action into phases, manually annotating their temporal boundaries. We note that each phase, let it be telic or atelic, is always delimited by a punctual action that denotes a transition from one phase to another. Therefore we represent durative actions as a sequence of telic or atelic phases, separated by punctual actions. An important advantage of exploiting punctual actions is that they are unambiguous and easy to annotate, since they define a precise instant in time in the development of the action (a single frame) and there is no need to specify their telicity or atelicity. This yields also to a a precise annotation of durative phases since punctual ones act as temporal boundaries. 

After identifying phases, we need to assign a progress value to each frame in an action tube. We base our phase-based progress formulation on the fact that each instance of a punctual phase should have the same progress value.
Since punctual phases act as boundaries for durative actions, which might exhibit high variability in duration and execution speed, this makes the overall resulting progress non linear.

To establish appropriate ground truth progress values, we first build a prototype action for each class, where the length of each durative phase is averaged across all instances in the training set. A generic instance of an action can be interpreted as a deformation of the prototype, where durative phases are compressed or dilated (i.e. slowed down or accelerated) with respect to their expected duration.
According to this, we label the action prototype with the linear interpretation of progress (Eq. \ref{eq:progress}), thus defining ground truth progress for punctual phases, since these are shared across instances.
Once punctual phases have been labeled, we can label durative phases knowing that their values must span over a well defined interval $[p_S, p_E]$ given by the progress of the two punctual boundaries.

To assign a progress value $p_i$ to frames belonging to durative phases, we adopt the following criteria, depending if the phase is telic or atelic:

\begin{itemize}
	\item $p_i = \frac{i}{N_i} + p_S$ for telic phases
	\item $p_i = \frac{(p_S + p_E)}{2}$ for atelic phases
\end{itemize}

where $i$ is the index of the frame within the current phase, $N_i$ the prototype duration of the current phase.
This corresponds to a linear progress for telic phases, which are usually executed uniformly since they represent events with a simple structure (rising hands, a jump, a kick of a ball). On the other hand, atelic phases can exhibit more complex behaviors with lack of structure. Since it is hard to establish a clear progression of states in atelic phases, we simply assign the expectation of progress inside the interval, based on the value of its boundaries.

It has to be noted that whereas the progress of prototype phases is defined linearly, the resulting progress values for real action tubes are in fact non linear (piecewise linear). Aligning the progress of each action tube in order to have the same value in correspondence of punctual phases is a form of action synchronization, where the execution rate of each phase is normalized in order to have the same speed. Executions of an action with a slow phase will therefore exhibit a slower growth of progress compared to other executions where the same phase happens quickly.

This definition of progress is more expressive than the linear interpretation, allowing us to model more complex dynamics.
To better understand the advantage of adopting the phase-based progress, in Fig. \ref{fig:progresses} a comparison is shown between the two types of annotations. When there is variability in the execution of a phase, progress values in the linear interpretation may become unaligned, yielding to an imprecise representation.

At the same time, the phase-based progress requires a demanding annotation process, since each action has to be divided into phases and all the boundaries need to be manually annotated. In Section \ref{sec:phase_annotation} the annotation procedure for the UCF-101 dataset is detailed. The collected annotations will be released upon publication.

\section{ProgressNet}\label{sec:model}

The whole architecture of our method is shown in Fig.~\ref{fig:arch}, highlighting the first branch dedicated to action classification and localization, and the second branch which predicts action progress. We believe that sequence modelling can have a huge impact on solving the task at hand, since time is a signal that carries a highly informative content.
Therefore, we treat videos as ordered sequences and propose a temporal model that encodes the action progress with a Recurrent Neural Network. In particular we use a model with two stacked Long Short-Term Memory layers (LSTM), with 64 and 32 hidden units respectively, plus a final fully connected layer with a sigmoid activation to predict action progress. Since actions can be also seen as transformations on the environment \cite{wang2016transf}, we feed the LSTMs with a feature representing regions and their context. We concatenate a contextual feature, computed by spatial pyramid pooling (SPP) of the whole frame \cite{he2014spatial}, with a region feature extracted with ROI Pooling \cite{ren2015faster}. The two representations are blended with a fully connected layer (FC7). The usage of a SPP layer allows us to encode contextual information for arbitrarily-sized images. We named this model \emph{ProgressNet}.

Our model emits a prediction $p_i \in [0,1] $ at each time step $i$ in an online fashion, with the LSTMs attention windows that keep track of the whole past history, \emph{i.e.}~from the beginning of the tube of interest until the current time step. We rely on an action detector~\cite{saha2016deep} to obtain scored boxes at each frame.
Such frame-wise action detectors are derived from object detector frameworks and fine-tuned on action bounding boxes. Features are extracted from the last convolutional feature map of such models by re-projecting each linked box onto it through ROI Pooling. Each tube is evaluated online independently in parallel.


We use ReLU non linearities after every fully connected layer, and dropout to moderate overfitting.  Our approach is \textit{online} and does not require complete tubes to perform predictions at test time. The only requirement of ProgressNet is to obtain a sequence of linked bounding boxes forming the tube.
Both online and offline solutions to this problem have been proposed \cite{saha2016deep,singh2017online}.
It has also to be noticed that ProgressNet adds a computational footprint of about 1ms per frame on a TITAN XP Pascal GPU, making it feasible to work in real time.
%

\subsection{Learning}\label{sec:training}
We initialize the spatio-temporal localization branch of our network, highlighted in orange in Fig.~\ref{fig:arch}, using a pre-trained action detector such as \cite{saha2016deep,singh2017online} while the remaining layers are learned from scratch. To train our ProgressNet we use ground truth action tubes as training samples.

To avoid overfitting the network, we apply the two following augmentation strategies. First, for every tube we randomly pick a starting point and a duration, so as to generate a shorter or equal tube (keeping the same ground truth progress values in the chosen interval). 
For instance, if the picked starting point is at the middle of the video and the duration is half the video length, the corresponding ground truth progress targets would be the values in the interval $[0.5, 1.0]$. This also forces the model to not assume 0 as starting progress value. 
Second, for every tube we generate a subsampled version, reducing the frame rate by a random factor uniformly distributed in $[1,10]$. This second strategy helps in generalizing with respect to the speed of execution of different instances of the same action class.

To encourage the network to be more precise on temporal boundaries of durative phases, we introduce a Boundary Observant (BO) loss. 
The idea is that phase boundaries (i.e. the punctual actions) do not have ambiguity and less error should be allowed on them. On the contrary, intermediate parts are less certain and a higher error can be tolerated.
We present our BO loss in a general form, considering actions composed by multiple phases. The linear interpretation is a simpler case of phase-based progress with a single durative phase, delimited by the start and the end of the whole action.

Given a prediction $\hat{p_i}$ and a target value $p_i$, for each phase $k$ of an action spanning across an interval $[l_k, u_k]$, we define a potential $e_k(p_i, \hat{p_i})$ with unitary cost for predictions that exceed the boundaries, decreasing it towards zero as the target gets closer to the center of the interval $m_k=(l_k + u_k)/2$:

\begin{equation}
e_k(p_i, \hat{p_i}) = \min \left [ 1, \left( \frac{p_{i} - m_k}{r_k\sqrt{2}} \right )^2 + \left( \frac{\widehat{p}_{i} - m_k}{r_k\sqrt{2}} \right )^2 \right ]
\end{equation}

The potential is derived from the circle equation $(x-x_c)^2 + (y-y_c)^2 = r^2$ of center $(x_c, y_c)$ and radius $r$. Since we want $e_k=1$ on the boundaries, we scale the circle to be circumscribed to the square of side $u_k - l_k$, i.e. the circle with radius $r_k \sqrt{2}$ centered in $m_k$ where $r_k=(u_k - l_k)/2$.

The potentials are computed for each phase and pooled together by a minimum operator. The Boundary Observant loss is then obtained by weighing the error with an L1 loss and averaging across samples:

\begin{equation}
\label{loss}
\mathcal{L}_{BO} = \frac{1}{N}\sum_{i}^{N} \min_k \left \{ e_k \right \} \left | p_{i} - \widehat{p}_{i} \right| 
\end{equation}

%
\begin{figure}[!t]
	\centering
	\begin{tabular}{ccc}
		\multirow{-4}{*}{\includegraphics[width=.3\columnwidth]{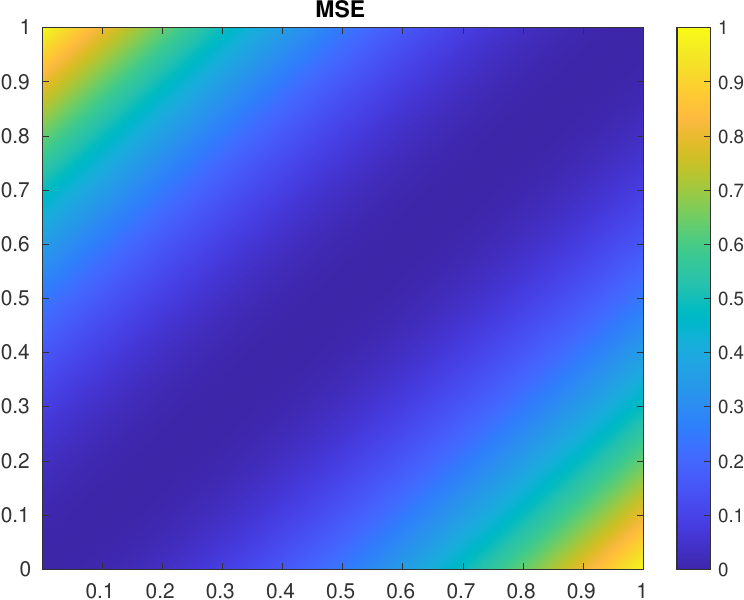}} &
		  \includegraphics[width=.28\columnwidth, trim={0 0 40pt 0}, clip]{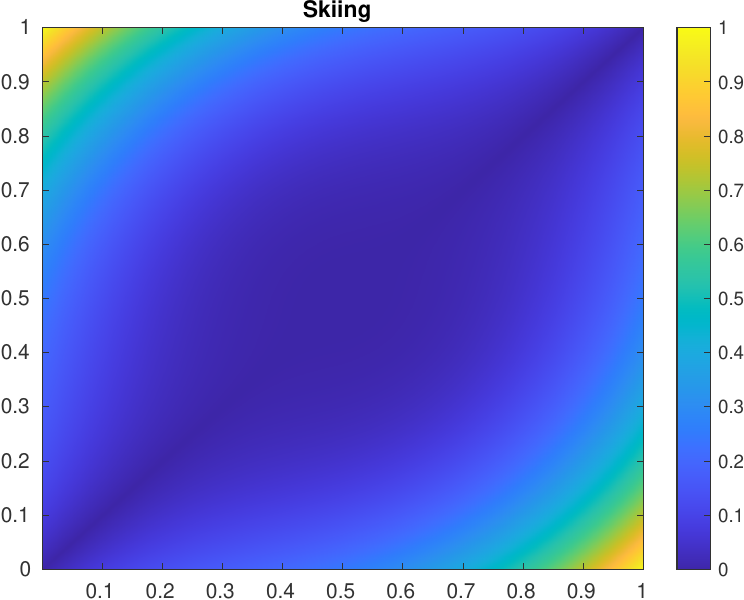}
		& \includegraphics[width=.28\columnwidth, trim={0 0 40pt 0}, clip]{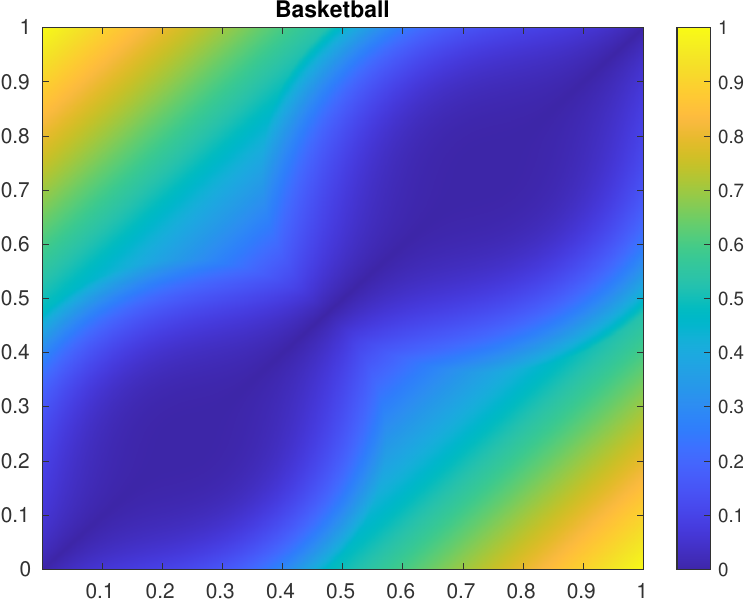} \\
		& \includegraphics[width=.28\columnwidth, trim={0 0 40pt 0}, clip]{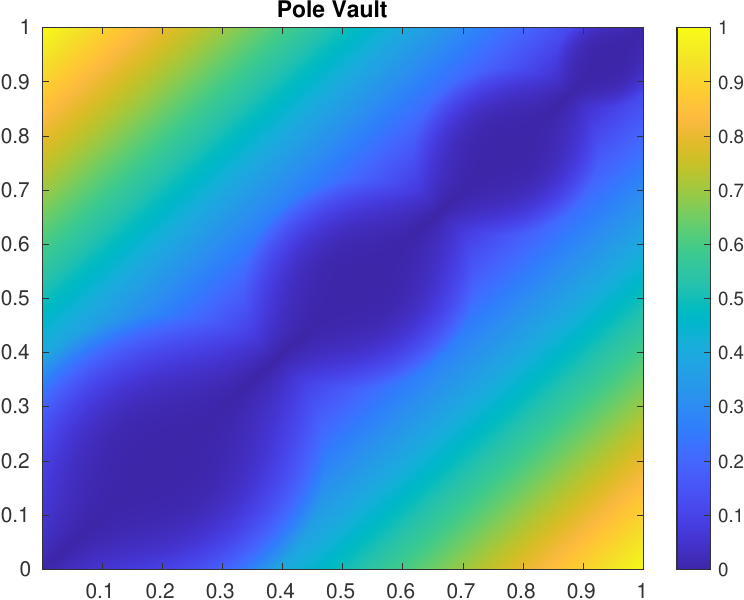}
		& \includegraphics[width=.28\columnwidth, trim={0 0 40pt 0}, clip]{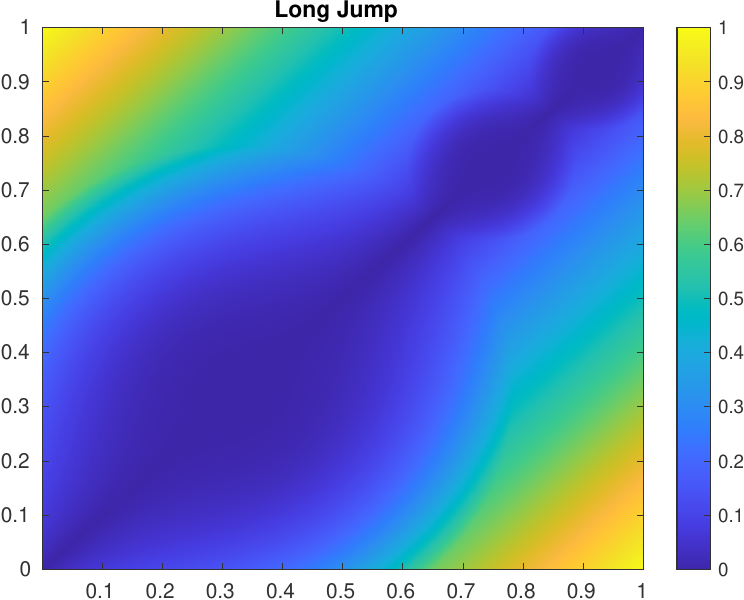}
	\end{tabular}
	\caption{Comparison between the $L2$ (left) and the Boundary Observant (right) loss functions. Predicted values and ground truth targets are on the two axes. It can be seen that the Boundary Observant loss is stricter against errors on the action boundaries. The Boundary Observant loss depends on the structure of each action, penalizing errors close phase boundaries. Examples for a few action classes are shown.}
	\label{fig:losses}
\end{figure}
Compared to the $L2$ loss for regression, the BO loss penalizes errors on the boundaries more than in intermediate parts, since we want to precisely identify when the phase starts and ends. At the same time, it avoids the trivial solution of always predicting the intermediate value $\widehat{p}=0.5$. Fig.~\ref{fig:losses} shows the difference between the two loss functions, where predicted values are on the $x$ axis and targets on the $y$ axis. Whereas the L2 loss function is always the same, BO adapts to the structure of the action and its phases. Note from the definition of progress that only values in $[0,1]$ can be expected.

We initialize all layers of ProgressNet with the Xavier \cite{glorot2010understanding} method and employ the Adam \cite{kingma2014adam} optimizer with a learning rate of $10^{-4}$. We use dropout with a probability of 0.5 on the fully connected layers.

\begin{table*}[t]
	\resizebox{.99\textwidth}{!}{
		\begin{tabular}{l||l|l|l|l|l|l|l|l|l|l|l}
			\textbf{Action}            & \textbf{Punctual}                     & \textbf{Durative}                             & \textbf{Punctual}                             & \textbf{Durative}                              & \textbf{Punctual}                             & \textbf{Durative}                             & \textbf{Punctual}                            & \textbf{Durative}                           & \textbf{Punctual}                             & \textbf{Durative}                        & \textbf{Punctual}                   \\ \hline
			\textit{Basketball}        & {\color[HTML]{\punctualcolor} Start} & {\color[HTML]{\teliccolor} Rising ball}   & {\color[HTML]{\punctualcolor} Shoot ball}    & {\color[HTML]{\teliccolor} Lowering hands} & {\color[HTML]{\punctualcolor} End}           &                                      & {\color[HTML]{\punctualcolor} }             &                                    & {\color[HTML]{\punctualcolor} }              &                                 & {\color[HTML]{\punctualcolor} }    \\
			\textit{BasketballDunk}    & {\color[HTML]{\punctualcolor} Start} & {\color[HTML]{\teliccolor} Jumping}       & {\color[HTML]{\punctualcolor} Dunk}          & {\color[HTML]{\teliccolor} Landing}        & {\color[HTML]{\punctualcolor} End}          &                                      & {\color[HTML]{\punctualcolor} }             &                                    & {\color[HTML]{\punctualcolor} }              &                                 & {\color[HTML]{\punctualcolor} }    \\
			\textit{Biking}            & {\color[HTML]{\punctualcolor} Start} & {\color[HTML]{\ateliccolor} Cycling}       & {\color[HTML]{\punctualcolor} End}           &                                       & {\color[HTML]{\punctualcolor} }              &                                      & {\color[HTML]{\punctualcolor} }             &                                    & {\color[HTML]{\punctualcolor} }              &                                 & {\color[HTML]{\punctualcolor} }    \\
			\textit{CliffDiving}       & {\color[HTML]{\punctualcolor} Start} & {\color[HTML]{\ateliccolor} Waiting}       & {\color[HTML]{\punctualcolor} Start jump}          & {\color[HTML]{\teliccolor} Diving}         & {\color[HTML]{\punctualcolor} Land in water} & {\color[HTML]{\ateliccolor} Swimming}      & {\color[HTML]{\punctualcolor} End}          &                                    & {\color[HTML]{\punctualcolor} }              &                                 & {\color[HTML]{\punctualcolor} }    \\
			\textit{CricketBowling}    & {\color[HTML]{\punctualcolor} Start} & {\color[HTML]{\ateliccolor} Running}       & {\color[HTML]{\punctualcolor} Start charge}  & {\color[HTML]{\teliccolor} Charging shot}  & {\color[HTML]{\punctualcolor} Shoot ball}    & {\color[HTML]{\teliccolor} Stop running}  & {\color[HTML]{\punctualcolor} End}          &                                    & {\color[HTML]{\punctualcolor} }              &                                 & {\color[HTML]{\punctualcolor} }    \\
			\textit{Diving}            & {\color[HTML]{\punctualcolor} Start} & {\color[HTML]{\ateliccolor} Approaching}   & {\color[HTML]{\punctualcolor} Start jump}    & {\color[HTML]{\teliccolor} Jumping}        & {\color[HTML]{\punctualcolor} Dive}          & {\color[HTML]{\ateliccolor} Somersaulting} & {\color[HTML]{\punctualcolor} Stretch body} & {\color[HTML]{\teliccolor} Falling}     & {\color[HTML]{\punctualcolor} Land in water} & {\color[HTML]{\ateliccolor} Swimming} & {\color[HTML]{\punctualcolor} End} \\
			\textit{Fencing}           & {\color[HTML]{\punctualcolor} Start} & {\color[HTML]{\ateliccolor} Fencing}       & {\color[HTML]{\punctualcolor} End}           &                                       & {\color[HTML]{\punctualcolor} }              &                                      & {\color[HTML]{\punctualcolor} }             &                                    & {\color[HTML]{\punctualcolor} }              &                                 & {\color[HTML]{\punctualcolor} }    \\
			\textit{FloorGymnastics}   & {\color[HTML]{\punctualcolor} Start} & {\color[HTML]{\ateliccolor} Standing}      & {\color[HTML]{\punctualcolor} Start running} & {\color[HTML]{\ateliccolor} Running}        & {\color[HTML]{\punctualcolor} Jump}          & {\color[HTML]{\ateliccolor} Somersaulting} & {\color[HTML]{\punctualcolor} Land}         & {\color[HTML]{\teliccolor} Rising arms} & {\color[HTML]{\punctualcolor} Arms up}       & {\color[HTML]{\ateliccolor} Standing} & {\color[HTML]{\punctualcolor} End} \\
			\textit{GolfSwing}         & {\color[HTML]{\punctualcolor} Start} & {\color[HTML]{\ateliccolor} Standing}      & {\color[HTML]{\punctualcolor} Start charge}  & {\color[HTML]{\teliccolor} Charging}       & {\color[HTML]{\punctualcolor} Max charge}    & {\color[HTML]{\teliccolor} Shooting}      & {\color[HTML]{\punctualcolor} Hit ball}     & {\color[HTML]{\teliccolor} Rising club} & {\color[HTML]{\punctualcolor} Top elevation} & {\color[HTML]{\ateliccolor} Waiting}  & {\color[HTML]{\punctualcolor} End} \\
			\textit{HorseRiding}       & {\color[HTML]{\punctualcolor} Start} & {\color[HTML]{\ateliccolor} Riding}        & {\color[HTML]{\punctualcolor} End}           &                                       & {\color[HTML]{\punctualcolor} }              &                                      & {\color[HTML]{\punctualcolor} }             &                                    & {\color[HTML]{\punctualcolor} }              &                                 & {\color[HTML]{\punctualcolor} }    \\
			\textit{IceDancing}        & {\color[HTML]{\punctualcolor} Start} & {\color[HTML]{\ateliccolor} Dancing}       & {\color[HTML]{\punctualcolor} End}           &                                       & {\color[HTML]{\punctualcolor} }              &                                      & {\color[HTML]{\punctualcolor} }             &                                    & {\color[HTML]{\punctualcolor} }              &                                 & {\color[HTML]{\punctualcolor} }    \\
			\textit{LongJump}          & {\color[HTML]{\punctualcolor} Start} & {\color[HTML]{\ateliccolor} Running}       & {\color[HTML]{\punctualcolor} Jump}          & {\color[HTML]{\teliccolor} Jumping}        & {\color[HTML]{\punctualcolor} Land}          & {\color[HTML]{\ateliccolor} Standing up}   & {\color[HTML]{\punctualcolor} End}          &                                    & {\color[HTML]{\punctualcolor} }              &                                 & {\color[HTML]{\punctualcolor} }    \\
			\textit{PoleVault}         & {\color[HTML]{\punctualcolor} Start} & {\color[HTML]{\ateliccolor} Running}       & {\color[HTML]{\punctualcolor} Pole down}     & {\color[HTML]{\teliccolor} Jumping}        & {\color[HTML]{\punctualcolor} Over bar}      & {\color[HTML]{\teliccolor} Falling}       & {\color[HTML]{\punctualcolor} End jump}     & {\color[HTML]{\ateliccolor} Standing up} & {\color[HTML]{\punctualcolor} End}           &                                 & {\color[HTML]{\punctualcolor} }    \\
			\textit{RopeClimbing}      & {\color[HTML]{\punctualcolor} Start} & {\color[HTML]{\ateliccolor} Climbing}      & {\color[HTML]{\punctualcolor} End}           &                                       & {\color[HTML]{\punctualcolor} }              &                                      & {\color[HTML]{\punctualcolor} }             &                                    & {\color[HTML]{\punctualcolor} }              &                                 & {\color[HTML]{\punctualcolor} }    \\
			\textit{SalsaSpin}         & {\color[HTML]{\punctualcolor} Start} & {\color[HTML]{\ateliccolor} Dancing}       & {\color[HTML]{\punctualcolor} End}           &                                       & {\color[HTML]{\punctualcolor} }              &                                      & {\color[HTML]{\punctualcolor} }             &                                    & {\color[HTML]{\punctualcolor} }              &                                 & {\color[HTML]{\punctualcolor} }    \\
			\textit{SkateBoarding}     & {\color[HTML]{\punctualcolor} Start} & {\color[HTML]{\ateliccolor} Skateboarding} & {\color[HTML]{\punctualcolor} End}           &                                       & {\color[HTML]{\punctualcolor} }              &                                      & {\color[HTML]{\punctualcolor} }             &                                    & {\color[HTML]{\punctualcolor} }              &                                 & {\color[HTML]{\punctualcolor} }    \\
			\textit{Skiing}            & {\color[HTML]{\punctualcolor} Start} & {\color[HTML]{\ateliccolor} Skiing}        & {\color[HTML]{\punctualcolor} End}           &                                       & {\color[HTML]{\punctualcolor} }              &                                      & {\color[HTML]{\punctualcolor} }             &                                    & {\color[HTML]{\punctualcolor} }              &                                 & {\color[HTML]{\punctualcolor} }    \\
			\textit{Skijet}            & {\color[HTML]{\punctualcolor} Start} & {\color[HTML]{\ateliccolor} Skijet}        & {\color[HTML]{\punctualcolor} End}           &                                       & {\color[HTML]{\punctualcolor} }              &                                      & {\color[HTML]{\punctualcolor} }             &                                    & {\color[HTML]{\punctualcolor} }              &                                 & {\color[HTML]{\punctualcolor} }    \\
			\textit{SoccerJuggling}    & {\color[HTML]{\punctualcolor} Start} & {\color[HTML]{\ateliccolor} Juggling}      & {\color[HTML]{\punctualcolor} End}           &                                       & {\color[HTML]{\punctualcolor} }              &                                      & {\color[HTML]{\punctualcolor} }             &                                    & {\color[HTML]{\punctualcolor} }              &                                 & {\color[HTML]{\punctualcolor} }    \\
			\textit{Surfing}           & {\color[HTML]{\punctualcolor} Start} & {\color[HTML]{\ateliccolor} Surfing}       & {\color[HTML]{\punctualcolor} End}           &                                       & {\color[HTML]{\punctualcolor} }              &                                      & {\color[HTML]{\punctualcolor} }             &                                    & {\color[HTML]{\punctualcolor} }              &                                 & {\color[HTML]{\punctualcolor} }    \\
			\textit{TennisSwing}       & {\color[HTML]{\punctualcolor} Start} & {\color[HTML]{\teliccolor} Opening}       & {\color[HTML]{\punctualcolor} Open}        & {\color[HTML]{\teliccolor} Hitting ball}   & {\color[HTML]{\punctualcolor} Hit ball}      & {\color[HTML]{\teliccolor} Closing}       & {\color[HTML]{\punctualcolor} Closed}       & {\color[HTML]{\ateliccolor} Moving}      & {\color[HTML]{\punctualcolor} End}           &                                 & {\color[HTML]{\punctualcolor} }    \\
			\textit{TrampolineJumping} & {\color[HTML]{\punctualcolor} Start} & {\color[HTML]{\ateliccolor} Jumping}       & {\color[HTML]{\punctualcolor} End}           &                                       & {\color[HTML]{\punctualcolor} }              &                                      & {\color[HTML]{\punctualcolor} }             &                                    & {\color[HTML]{\punctualcolor} }              &                                 & {\color[HTML]{\punctualcolor} }    \\
			\textit{VolleyballSpiking} & {\color[HTML]{\punctualcolor} Start} & {\color[HTML]{\ateliccolor} Running}       & {\color[HTML]{\punctualcolor} Start jump}    & {\color[HTML]{\teliccolor} Jumping}        & {\color[HTML]{\punctualcolor} Hit ball}      & {\color[HTML]{\teliccolor} Landing}       & {\color[HTML]{\punctualcolor} Land}         & {\color[HTML]{\ateliccolor} Moving}      & {\color[HTML]{\punctualcolor} End}           &                                 & {\color[HTML]{\punctualcolor} }    \\
			\textit{WalkingWithDog}    & {\color[HTML]{\punctualcolor} Start} & {\color[HTML]{\ateliccolor} Walking}       & {\color[HTML]{\punctualcolor} End}           &                                       & {\color[HTML]{\punctualcolor} }              &                                      & {\color[HTML]{\punctualcolor} }             &                                    & {\color[HTML]{\punctualcolor} }              &                                 & {\color[HTML]{\punctualcolor} }                             
			
		\end{tabular}
	}
	\caption{Action phases annotated for the UCF-101 dataset. Punctual phases are denoted in {\color[HTML]{\punctualcolor} blue}, telic in {\color[HTML]{\teliccolor} green} and atelic in {\color[HTML]{\ateliccolor} orange}.}
	\label{tab:phases}
\end{table*}

\section{Experimental Setting}\label{sec:experimental_setting}

In this section we discuss the experimental setting to evaluate action progress prediction for spatio-temporal tubes and propose two evaluation protocols. We introduce some simple baselines and show the benefits of our approach.

We experiment on the J-HMDB \cite{jhuang2013towards} and UCF-101 \cite{soomro2012ucf101} datasets. J-HMDB consists of 21 action classes and 928 videos, annotated with body joints from which spatial boxes can be inferred. All the videos are temporally trimmed and contain only one action. Since clips are all very short, all actions can be considered to be telic. We use this dataset to benchmark action progress prediction with a linear interpretation.
UCF-101 contains 24 classes annotated for spatio-temporal action localization. It is a more challenging dataset because actions are temporally untrimmed and there can be more than one action of the same class per video. Moreover, it contains video sequences with large variation in appearance, scale and illumination.
We use this dataset to predict action progress both with the linear and the phase-based interpretations of progress, which required to manually identify and annotate all sub-phases of the actions in the dataset. Section \ref{sec:phase_annotation} details the annotation process and which phases have been used.
We adopt the same split of UCF-101 used in~\cite{weinzaepfel2015learning,peng2016multi,saha2016deep,yu2015fast}.

Note that larger datasets such as THUMOS~\cite{idrees2017thumos} and ActivityNet~\cite{caba2015activitynet} do not provide bounding box annotations, and therefore can not be used in our setting.

\subsection{Phase annotation}\label{sec:phase_annotation}

We annotated phases for all actions in the UCF-101 dataset~\cite{soomro2012ucf101}.
Punctual events are manually identified and used as boundaries for durative phases. We specify whether a durative phase is telic or atelic to assign the proper progress values as defined in Section \ref{sec:phasebased}.

We observe that the original annotations have imprecise temporal boundaries for our task. 
For instance, there is a different amount of waiting before a golf swing or a dive. 
We believe that such durative atelic phases are tied to the known difficulties \cite{moltisanti2017trespassing} in clearly identifying the temporal boundaries of an action where \textit{Pre-actional} phases may be present. 
Hence, by annotating punctual events, the annotator has unambiguous instructions about which frame to annotate and the uncertainty is reduced.
We split every action in the dataset as in Table \ref{tab:phases}. To easily identify phases and facilitate the annotation process, we use verbs in their \textit{-ing} form to denote durative phases and in the infinitive form for punctual ones.

The annotators were asked to identify punctual actions (i.e. select the corresponding frame) for each action tube, starting from the original spatio-temporal annotations\footnote{We used the revised annotations available at \url{https://github.com/gurkirt/corrected-UCF101-Annots}}. Durative phases are automatically labeled since they are defined by two punctual boundaries. We have identified a total of 27,120 phases, 15,789 of which are punctual and 11,331 durative.

\subsection{Metrics}
\label{evaluationmetrics}
In order to evaluate the task of action progress prediction, we introduce two evaluation metrics.

\paragraph{Framewise Mean Squared Error}
This metric tells how well the model behaves at predicting action progress when the spatio-temporal coordinates of the actions are known.
Test data is evaluated frame by frame by taking the predictions $\widehat{p}_{i}$ on the ground truth boxes $B_i$ and comparing them with action progress targets $p_{i}$. We compute mean squared error
$MSE =  || \widehat{p}_{i} - p_{i} ||^2$
across each class. Being computed on ground truth boxes, this metric assumes perfect detections and thus disregards the action detection task, only evaluating how well progress prediction works.

\paragraph{Average Progress Precision}
Average Progress Precision (APP) is identical to framewise Average Precision (Frame-AP) \cite{ggjm2015tubes} with the difference that true positives must have a progress that lays within a margin from the ground truth target.
Frame-AP measures the area under the precision-recall curve for the detections in each frame. A detection is considered a hit if its Intersection over Union (IoU) with the ground truth is bigger than a threshold $\tau$ and the class label is correct.
In our case, we fix $\tau=0.5$ and evaluate the results at different progress margins $m$ in $[0,1]$.
A predicted bounding box $\widehat{B}_i$ is matched with a ground truth box $B_i$ and considered a true positive when $B_i$ has not been already matched and the following conditions are met:
\begin{equation}
IoU(\widehat{B}_i, B_i) \geq \tau,  \quad |\widehat{p}_i - p_i | \leq m
\end{equation}

\noindent where $\widehat{p}_i$ is the predicted progress, $p_i$ is the ground truth progress and $m$ is the progress margin. We compute Average Progress Precision  for each class and report a mean (mAPP) value for a set of $m$ values.


\begin{figure*}[!t]
	\centering
	\includegraphics[width=.9\textwidth]{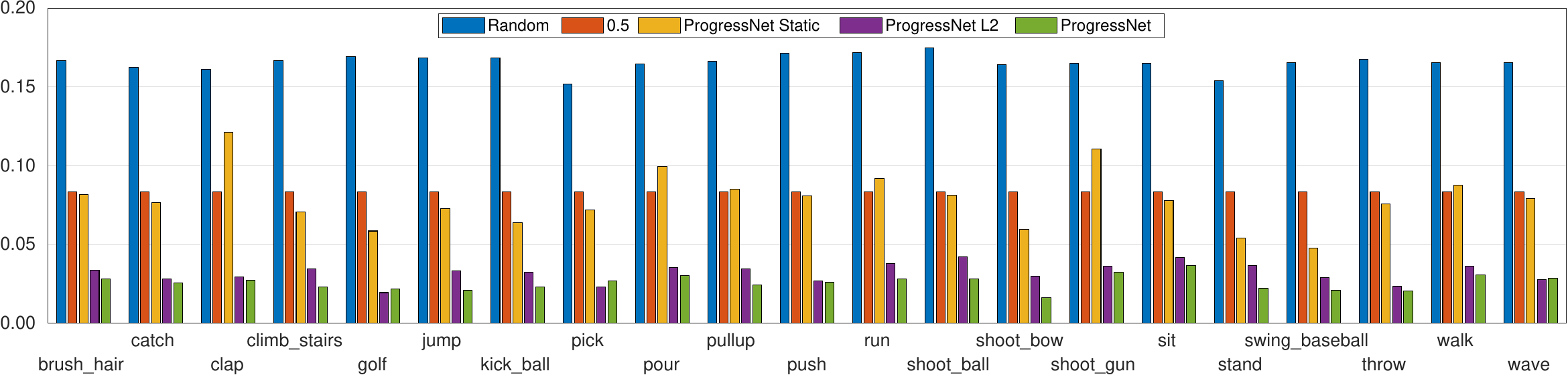}
	\caption{Mean Squared Error obtained by three formulations of our model (ProgressNet Static, ProgressNet L2 and ProgressNet) on the J-HMDB dataset. Random and constant 0.5 predictions are reported as reference.}
	\label{fig:mse_static_temporal_jhmdb}
\end{figure*}

\begin{figure*}[!t]
	\centering
	\begin{tabular}{@{}cc@{}}
		\begin{tabular}[t]{@{}cc@{}}
			\raisebox{-\height}{\includegraphics[width=0.22\textwidth]{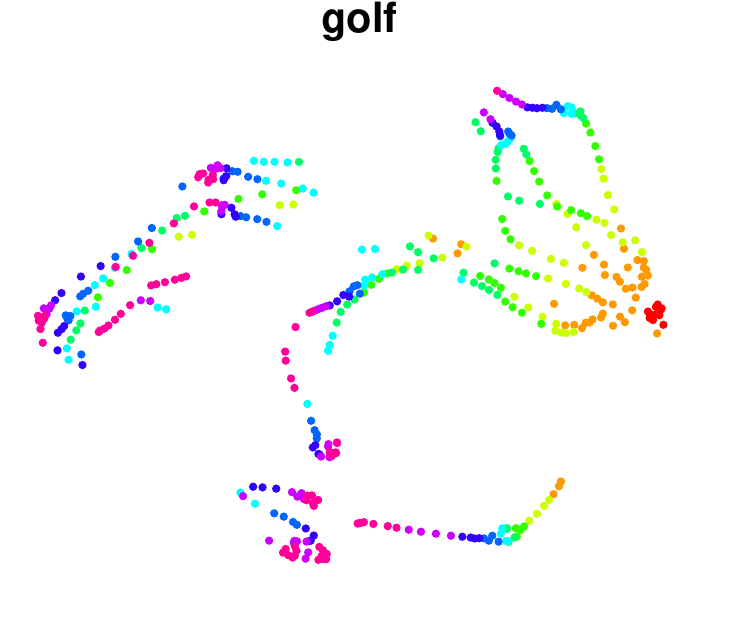}} & 
			\raisebox{-\height}{\includegraphics[width=0.22\textwidth]{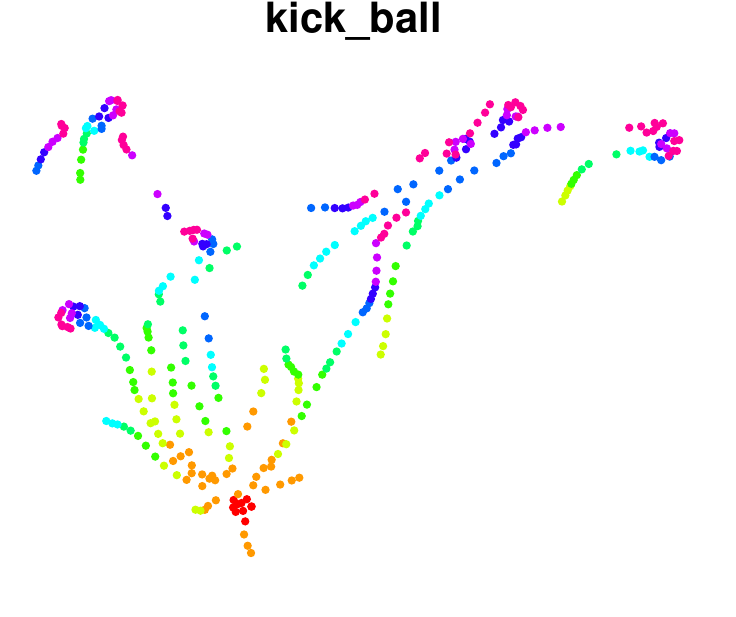}} \\ 
			\raisebox{-\height}{\includegraphics[width=0.22\textwidth]{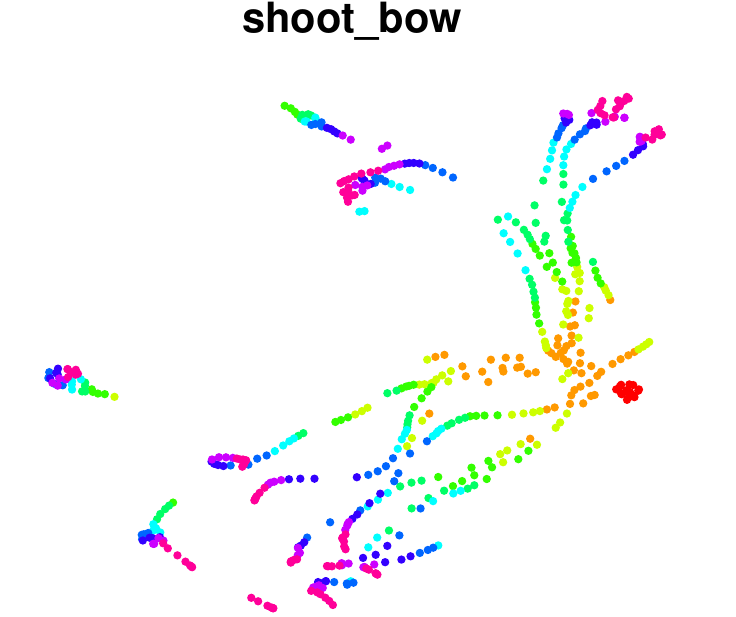}} & 
			\raisebox{-\height}{\includegraphics[width=0.22\textwidth]{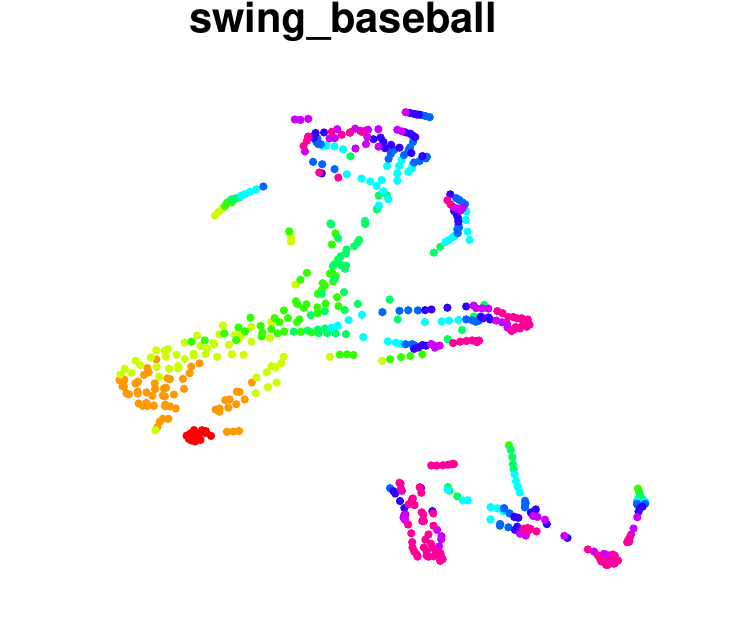}}
		\end{tabular} & 
		\raisebox{-\height}{\includegraphics[width=0.5\textwidth]{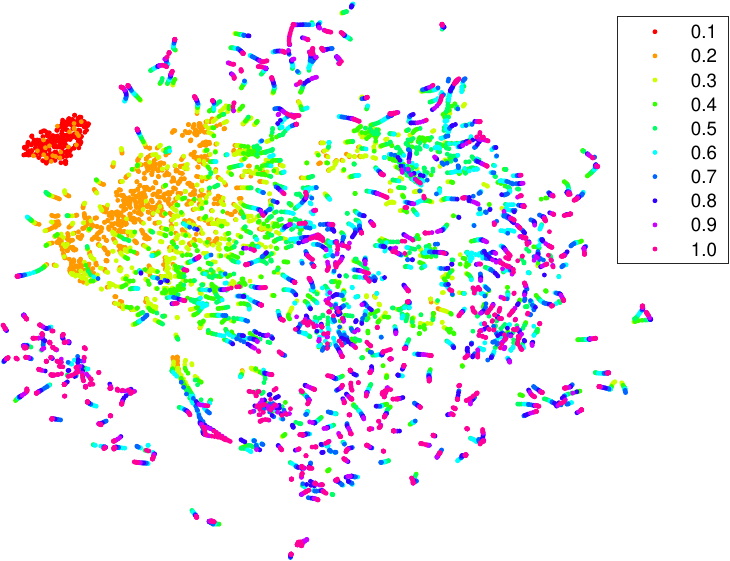}}
	\end{tabular}
	\caption{t-SNE visualizations of ProgressNet's hidden states of the second LSTM layer on the J-HMDB test set. Each point corresponds to a frame colored with its ground truth progress, quantized with a 0.1 granularity. On the left, the states of 4 classes are shown separately, while in the right all the test frames are shown together.}
	\label{fig:tsne-jhmdb-all}
\end{figure*}

\subsection{Implementation Details}

In practice, we observed that on some classes of the UCF-101 dataset it is hard to learn accurate progress prediction models.
These are action classes like \textit{Biking} or \textit{WalkingWithDog} that are completely atelic and even for a human observer is hard to guess how far the action has progressed.
Therefore we extended our framework by adopting a curriculum learning strategy. First, the model is trained as described in Sect.~\ref{sec:training} on classes that have at least a telic phase\footnote{This subset consists of the following classes: \textit{Basketball}, \textit{BasketballDunk}, \textit{CliffDiving}, \textit{CricketBowling}, \textit{Diving}, \textit{FloorGymnastics}, \textit{GolfSwing}, \textit{LongJump}, \textit{PoleVault}, \textit{TennisSwing}, \textit{VolleyballSpiking}.}.
Then, we fix all convolutional, FC and LSTM layers and fine-tune the FC8 layer that is used to perform progress prediction from the last LSTM output on the whole UCF-101 dataset. This strategy improves the convergence of the model.


\section{Experiments}

In this section we report the experimental results on the task of predicting action progress.
We first present an analysis of ProgressNet using a linear progress formulation and then using phase-based progress.
Since ProgressNet is a multitask approach, we start by measuring progress estimation on perfectly localized actions, i.e. discarding possible action localization errors. In addition we perform several ablation studies, underlining the importance of the Boundary Observant loss and the behavior of our method on partially observed action tubes.
We then test the method on real detected tubes with the full model and finally report a qualitative analysis which shows some success and failure cases.

\subsection{Linear Progress}

\paragraph{\textbf{Action progress on correctly localized actions}}
In this first experiment we evaluate the ability of our method to predict action progress on correctly localized actions in both time and space.
We take the ground truth tubes of actions on the test set and compare the MSE of three variants of our method: the full architecture trained with our Boundary Observant loss (\textit{ProgressNet}), the same model trained with L2 loss (\textit{ProgressNet L2}) and a reduced memoryless variant (\textit{ProgressNet Static}).

The comparison of our full model against ProgressNet L2 is useful to understand the contribution of the Boundary Observant loss with respect to a simpler L2 loss.
To underline the importance of using recurrent networks in action progress prediction, in the variant ProgressNet Static we substitute the two LSTMs with two fully connected layers that predict progress framewise.

In Tab.~\ref{tab:mse_dataset} we report the MSE results for a linear progress on both the J-HMDB and UCF-101 datasets. In addition to the variants of our models, we provide two baselines: random prediction and constant prediction.
The random prediction provides us with a higher bound on the MSE values. 
For the constant prediction, we always predict the progress expectation $\widehat{p}=0.5$ for every frame, which is a trivial solution that obtains good MSE results. Both are clearly far from being informative for the task. 

\begin{table}[!t]
	\centering
	\begin{tabular}{l|c|c}
		& {J-HMDB} & {UCF-101}   \\ \hline
		Random             & 0.166  & 0.166  \\ 
		0.5                & 0.084  & 0.083  \\ 
		ProgressNet Static & 0.079  & 0.104  \\ 
		ProgressNet L2     & 0.032  & 0.052  \\ 
		ProgressNet        & \textbf{0.026}  & \textbf{0.049}  \\ \hline
	\end{tabular}
	%
	\caption{Mean Square Error values for action progress prediction on the UCF-101 and J-HMDB datasets. Results are averaged among all classes.}
	\label{tab:mse_dataset}
\end{table}
 
We first observe that the MSE values are consistent among the two datasets, with ProgressNet models ahead of the other methods. ProgressNet and ProgressNet L2 obtain a much lower error than ProgressNet Static and the baselines. This confirms the ability of our model to understand action progress.
In particular, the best result is obtained with ProgressNet, proving that our Boundary Observant loss plays an important role in training the network effectively.
ProgressNet Static has an inferior MSE than the variants with memory, suggesting that single frames for some classes can be ambiguous and a temporal context can help to accurately predict action progress.
In particular, observing the class breakdown for J-HMDB in Fig.~\ref{fig:mse_static_temporal_jhmdb}, we note that the static model gives better MSE values for some actions such as \textit{Swing Baseball} and \textit{Stand}. 
This is due to the fact that such actions have clearly identifiable states, which help to recognize the development of the action. On the other hand, classes such as \textit{Clap} and \textit{Shoot Gun} are hardly addressed with models without memory because they exhibit only few key poses that can reliably establish the progress.

\begin{figure*}[!t]
	\centering
	\begin{tabular}{cc}
		\includegraphics[width=.9\textwidth]{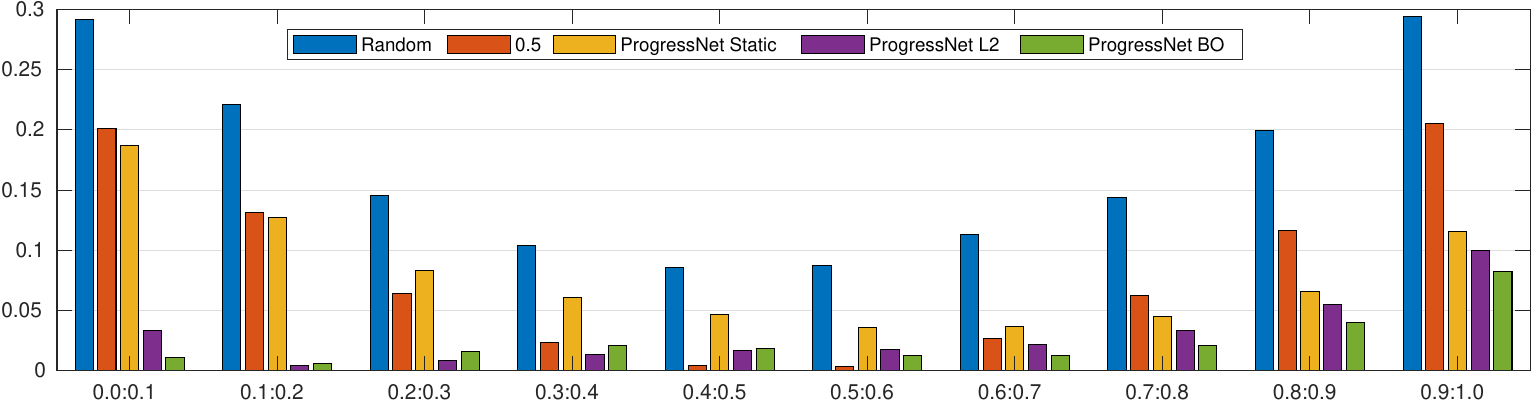}
	\end{tabular}
	\caption{MSE breakdown grouped by progress intervals for the J-HMDB dataset. Similar results are obtained with UCF-101.}
	\label{fig:mse_histogram}
\end{figure*}

In Fig.~\ref{fig:tsne-jhmdb-all} we show t-SNE~\cite{maaten2008visualizing} embeddings of the hidden states of the second LSTM layer. Each point is a frame of the test set of J-HMDB and is colored according to its true action progress. We report both the embeddings for the whole test set and for 4 classes separately. In all figures we note that progress increases radially along trajectories from points labeled with 0.1. This suggests that ProgressNet has learned directions in the hidden state space that follow action progress.

\paragraph{\textbf{Error at different progress points}}
To understand when our model is more prone to errors, with respect to the action progression, in Fig.~\ref{fig:mse_histogram} we show a breakdown of the error by dividing the MSE obtained in the previous experiment according to the ground truth progress. It can be seen how action progress is harder at the boundaries and how our Boundary Observant loss helps in mitigating this difficulty.

\paragraph{\textbf{Expected length and partially observed tubes}}

Despite the simple progress model, our approach is not just predicting incrementing linear values over an expected tube length. To show that we are able to understand the action progress even on truncated or generally incomplete tubes, we test an additional baseline where progress is estimated as the ratio between the frame ID in the tube and the expected length of the predicted class (estimated on the training videos).

\begin{figure}[!h]
	\centering
	\includegraphics[width=\columnwidth]{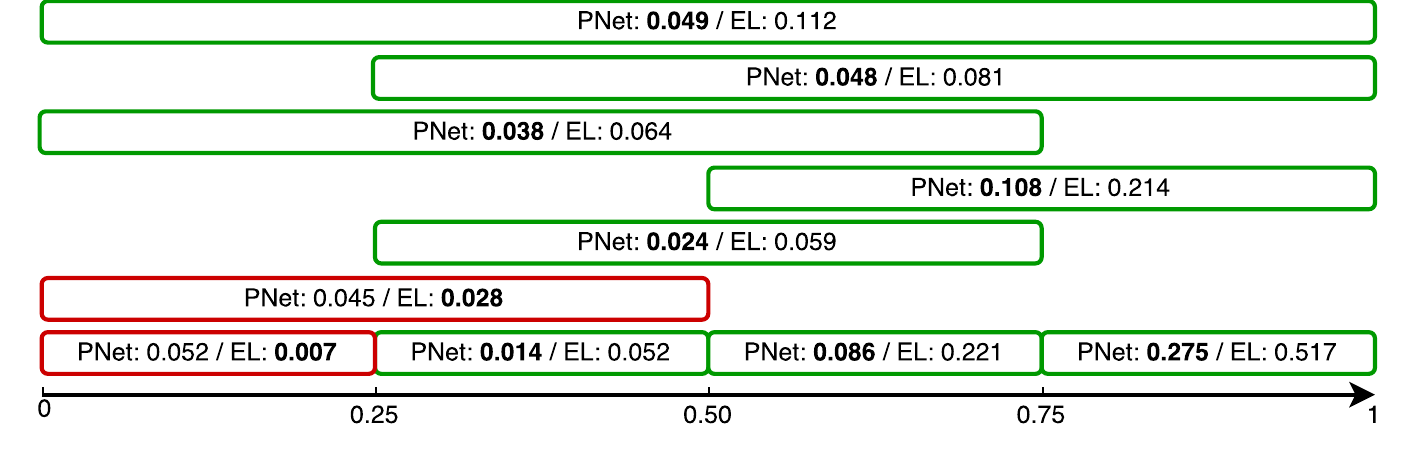}
	\caption{MSE on partially observed tubes of UCF-101. Each block depicts an experimental setting with a different tube fraction, picking the beginning and the end in $[0,0.25,0.5,0.75,1.0]$. The first value is the MSE of ProgressNet (PNet) while the second one of the expected length baseline (EL). ProgressNet wins in 8 out of 10 cases.}
	\label{fig:partialtubes}
\end{figure}

This baseline obtains on UCF-101 an MSE of 0.112, which is only lower than Random (0.166) and is largely outperformed by our method. Moreover we show how it suffers on partially observed tubes. In Fig.~\ref{fig:partialtubes} we report MSE values for ProgressNet and the expected length baseline varying the observation window of the tubes. ProgressNet largely outperforms the baseline except when observing a small portion of the tube at the beginning of the action.

\begin{figure}[!ht]
	\centering
	\begin{tabular}{cc}
		\includegraphics[width=0.48\columnwidth]{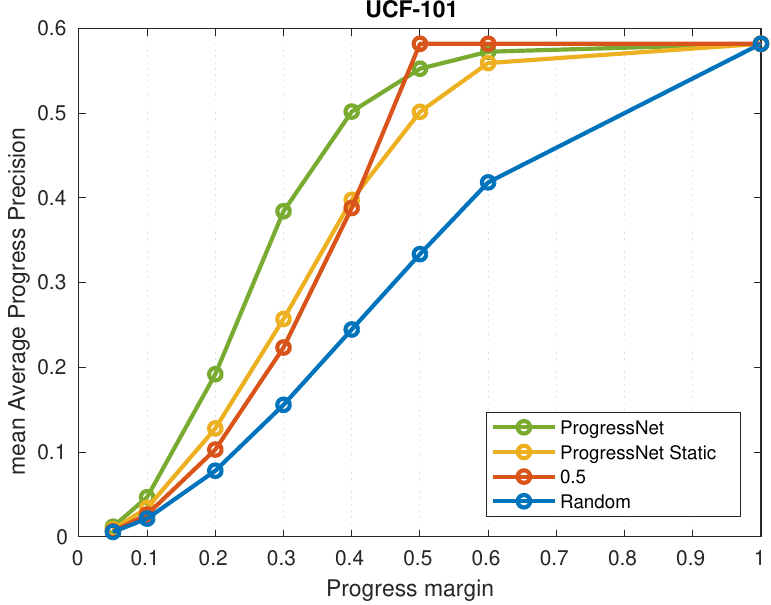}
		\includegraphics[width=0.48\columnwidth]{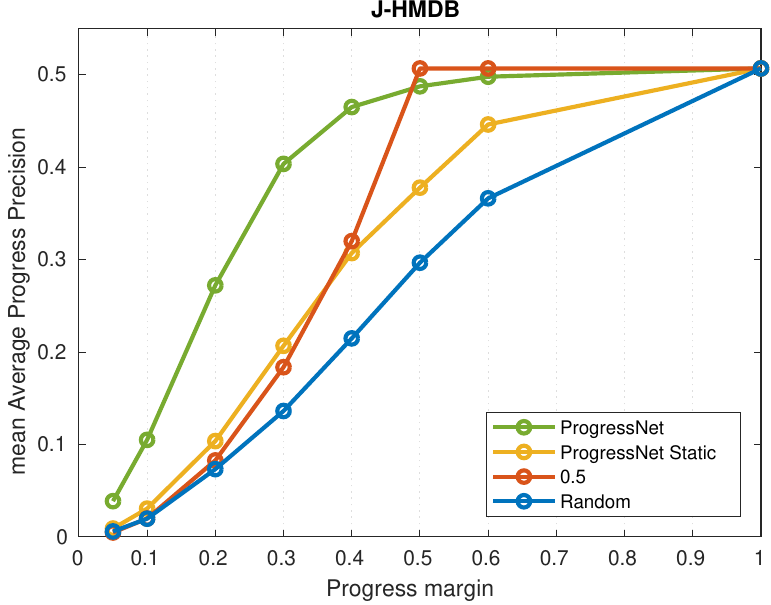}
		
	\end{tabular}
	\caption{mAPP on the UCF-101 and J-HMDB datasets.}
	\label{fig:mAPP_plots}
\end{figure}

\paragraph{\textbf{Action progress with the full pipeline}}
In this experiment, we evaluate the performance of action progress while also performing the spatio-temporal action detection with the entire pipeline. Differently from the previous experiment, we test the full approach where action tubes are generated by the detector.
In Fig.~\ref{fig:mAPP_plots}, we report the mAPP of ProgressNet (trained with BO loss), ProgressNet Static and the two baselines Random and $0.5$ on both the UCF-101 and J-HMDB benchmarks.
Note that the mAPP upper bound is given by standard mean Frame-AP \cite{ggjm2015tubes}, which is equal to mAPP with margin $m=1$. In the $\widehat{p}=0.5$ baseline, this upper bound is reached with $m=0.5$.

It can be seen that ProgressNet has mAPP higher than the baselines for stricter progress margin. This confirms that our approach is able to predict action progress correctly even when tubes are noisy such as those generated by a detector.  
ProgressNet Static exhibits a lower performance than ProgressNet, confirming again that the memory is helpful to model action progress. 

\paragraph{\textbf{Architecture of the Network}}
We trained ProgressNet with just the first LSTM layer and report an increase of MSE to 0.081 on the UCF-101 dataset. This performs on par with the 0.5 baseline (0.083) and providing good results only on actions with simple dynamics (i.e. \textit{Diving}, \textit{GolfSwing}).

\subsection{Phase-based progress}
We trained ProgressNet also using phase-based annotations on UCF-101. Labeling actions with a phase based progress, allows us to provide a better characterization of the evolution of the action.

\begin{figure*}[!t]
	\centering
	\includegraphics[width=.9\textwidth]{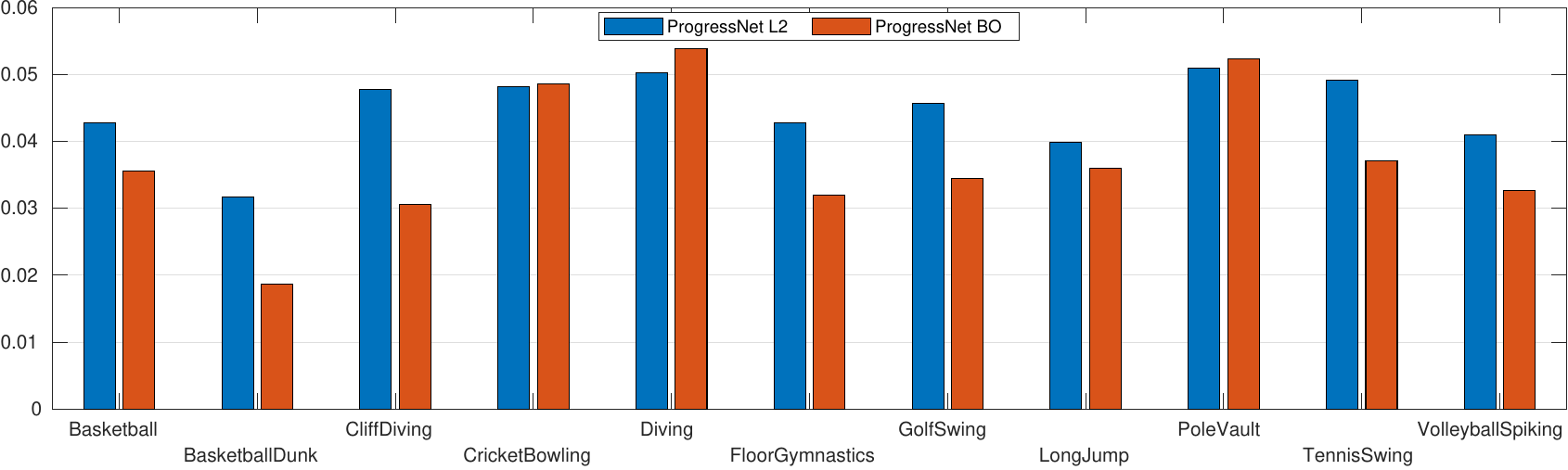}
	\caption{MSE values with phase-based progress for telic actions, comparing ProgressNet trained with the L2 and BO losses.}
	\label{fig:l2_vs_bo_phase}
\end{figure*}

\paragraph{\textbf{Action progress on correctly localized actions.}}
In Tab.~\ref{tab:mse_phase} we report the MSE obtained by the three variants of our method. Again, ProgressNet Static obtains a much higher error than the versions equipped with LSTMs and using the Boundary Observant loss slightly improves the capability of the network over an L2 loss. Interestingly, the biggest gain is observed on telic actions (i.e. actions with at least a telic phase), whereas atelic action do not seem to be affected by the BO loss. This is due to the fact that purely atelic actions do not have well defined boundaries. In fact in the case of UCF-101 the boundaries of atelic actions often correspond with starting and ending frames of the videos. To better underline the importance of the Boundary Observant loss for telic actions, a class-wise comparison of ProgressNet with L2 and BO losses is shown in Fig.~\ref{fig:l2_vs_bo_phase}. It can be seen that most actions improve considerably when the model is forced to perform well on boundaries.

\begin{table}[!t]
	\centering
	\begin{tabular}{l|c|c|c}
		& All & Telic & Atelic \\ \hline
		ProgressNet Static & 0.142 & 0.197 & 0.132 \\ 
		ProgressNet L2     & 0.024 & 0.045 & \textbf{0.008} \\ 
		ProgressNet        & \textbf{0.021} & \textbf{0.037} & 0.010 \\ \hline
	\end{tabular}
	\caption{Mean Squared Error values on UCF-101 with phase-based progress. Results are shown averaged over all classes as well as considering only telic actions or atelic actions.}
	\label{tab:mse_phase}
\end{table}

\begin{figure*}[!t]
	\centering
	\includegraphics[width=.99\textwidth]{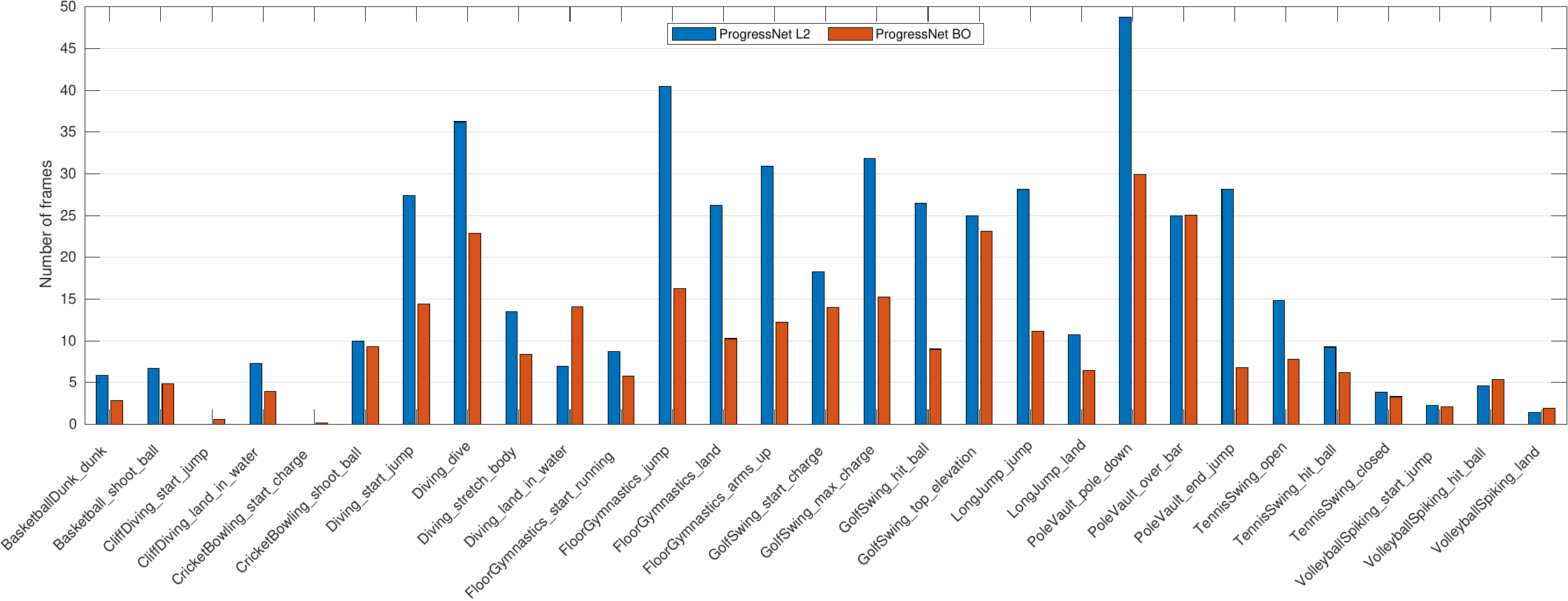}
	\caption{Temporal offset between real and predicted progress for punctual phases, measured in number of frames.}
	\label{fig:punctual_offset}
\end{figure*}

Since the Boundary Observant loss forces training to penalize errors more on phase boundaries, we investigate the behavior of the model on punctual phases. 
Since such phases are always defined within a single frame, we measure the average number of frames $\Delta_{f}$ between the frame $f_k$ representing the punctual phase and the frame with the closest predicted progress $\hat{p}$:
\begin{equation}
	\Delta_{f} = \frac{1}{K}\sum_{k=1}^{K}|f_k - \arg\min_i (p_k - \hat{p}_i)|
\end{equation}
This metric highlights the temporal offset between the punctual phase and when its progress is estimated by the model. Also in this case, as can be observed in Fig. \ref{fig:punctual_offset}, a comparison with the L2 loss highlights the importance of the Boundary Observant loss. The model largely benefits from the usage of the loss, with the temporal lag of the predictions $\Delta_{f}$ being reduced to under a second (25 frames) in almost all cases.


\begin{figure}[!t]
	\centering
	\begin{tabular}{cc}
		\includegraphics[width=0.5\columnwidth]{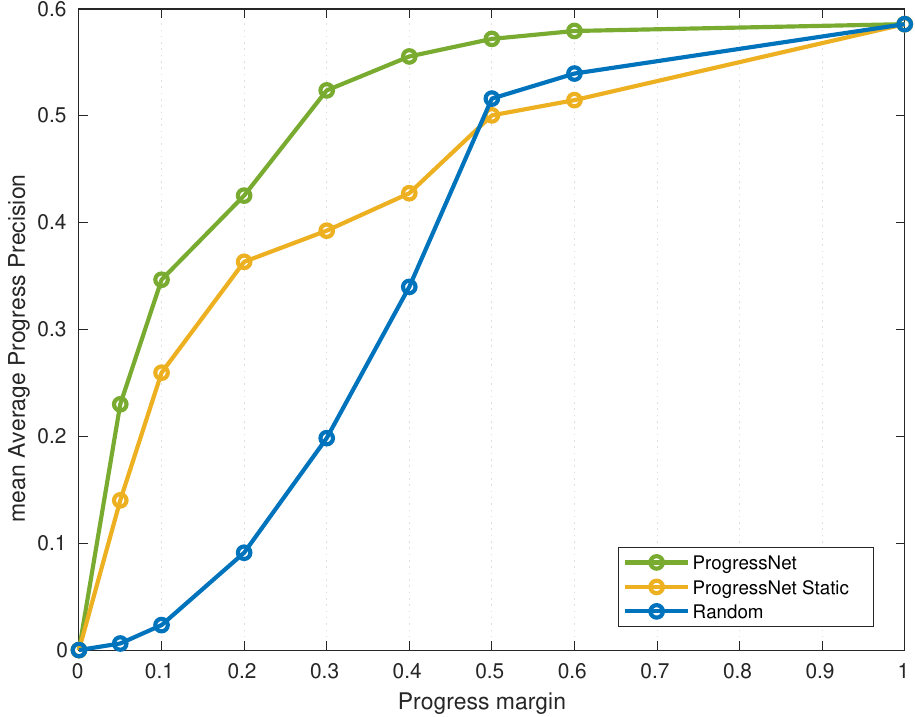}
	\end{tabular}
	\caption{mAPP on the UCF-101 using phase-based annotations.}
	\label{fig:mAPP_plot_phase}
\end{figure}

\paragraph{\textbf{Action progress with the full pipeline.}}
When testing ProgressNet with the full pipeline, i.e. evaluating also how well the localization branch performs in conjunction with progress estimation, we can see that predictions are closer to the ground truth compared to the same model trained with the linear progress formulation. This hints to the fact that phase-based progress reduces annotation ambiguity and therefore aids the optimization of the model.
This trend can be observed in Fig. \ref{fig:mAPP_plot_phase}, where the mAPP for ProgressNet and ProgressNet Static, along with the Random baseline, is reported varying the margin threshold. Interestingly, even the static version of the model is able to perform better than its counterpart with linear progress labels (see Fig. \ref{fig:mAPP_plots}), suggesting once again that the model is able to better understand the visual cues that are part of the development of the action.

\subsection{Qualitative analysis}
\begin{figure*}[t!]
	\bgroup
	\setlength\tabcolsep{4pt}
	
	\begin{tcolorbox}[colframe=green, left=-4pt, right=0pt,top=0pt, bottom=-4pt]
		\centering

		\begin{tabular}{ccccc}
			\includegraphics[width=.18\textwidth, trim={0 42pt 0 42pt},clip] {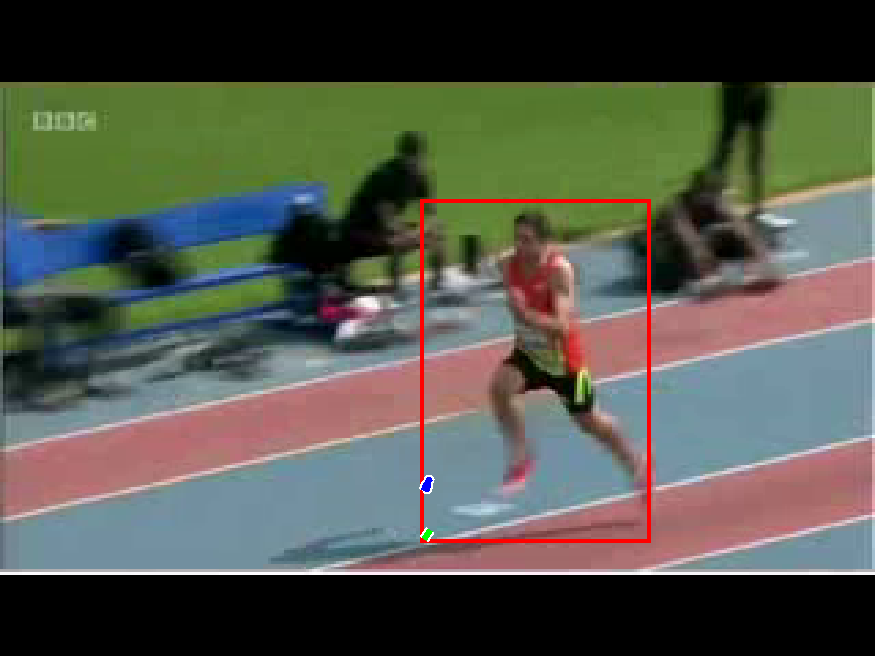}
			& \includegraphics[width=.18\textwidth, trim={0 42pt 0 42pt},clip] {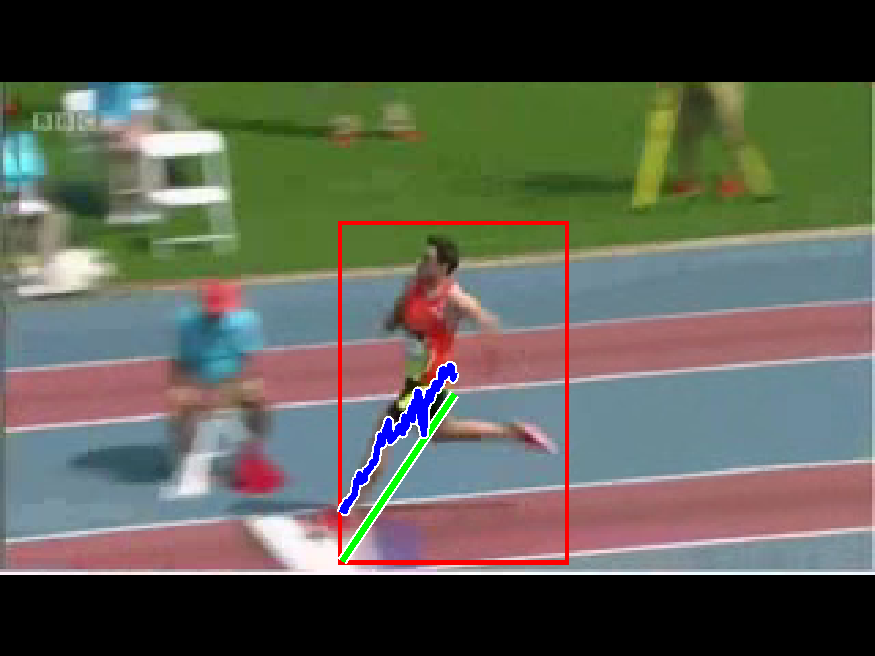}
			& \includegraphics[width=.18\textwidth, trim={0 42pt 0 42pt},clip] {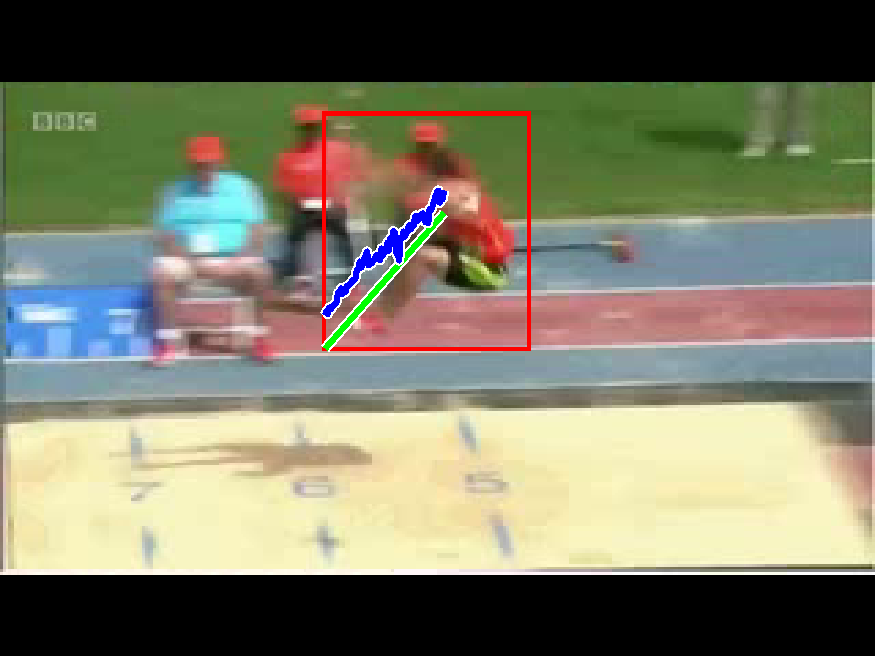}
			& \includegraphics[width=.18\textwidth, trim={0 42pt 0 42pt},clip] {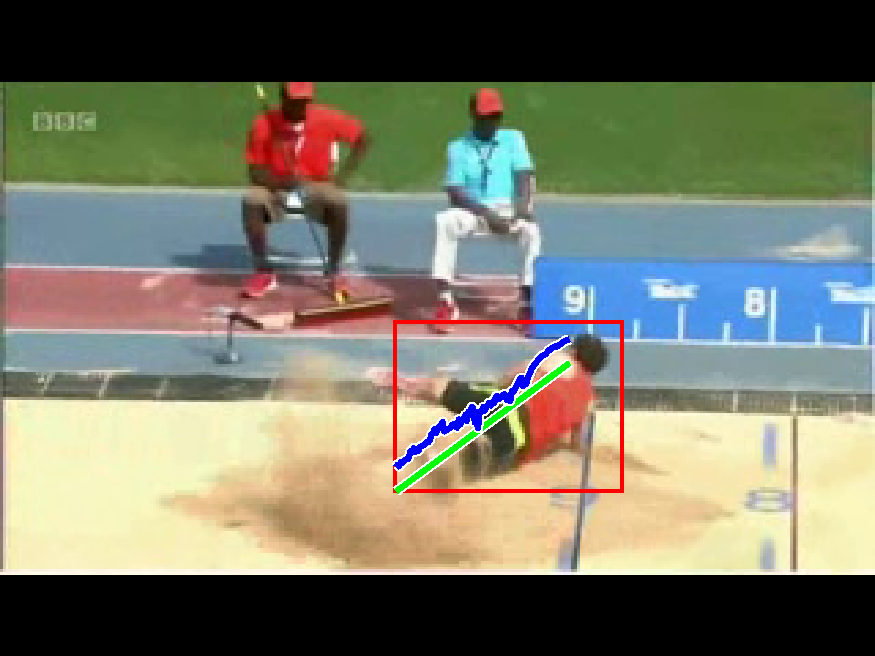}
			& \includegraphics[width=.18\textwidth, trim={0 42pt 0 42pt},clip] {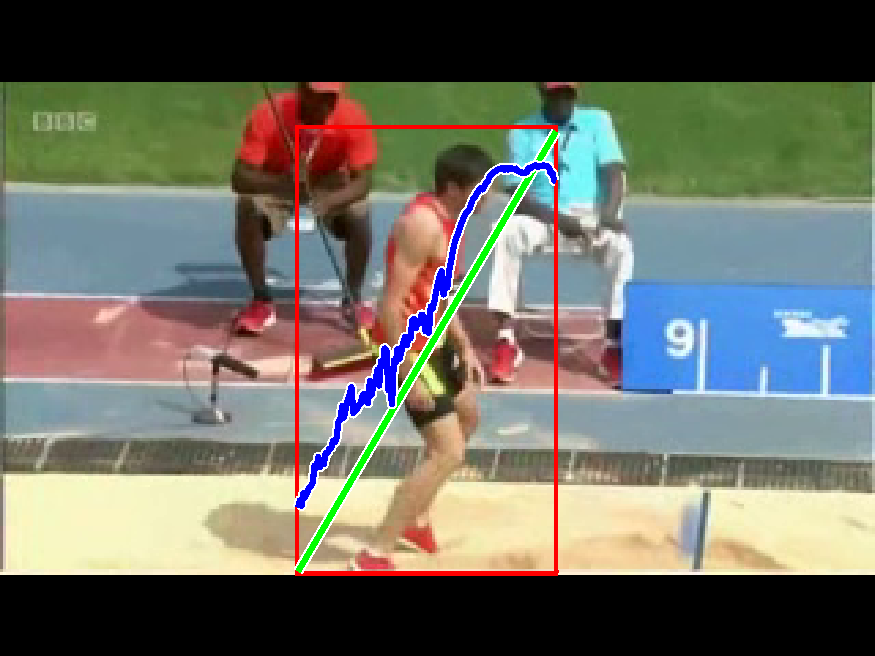}
		\end{tabular}
		
		\begin{tabular}{ccccc}
			\includegraphics[width=.18\textwidth, trim={0 42pt 0 42pt},clip] {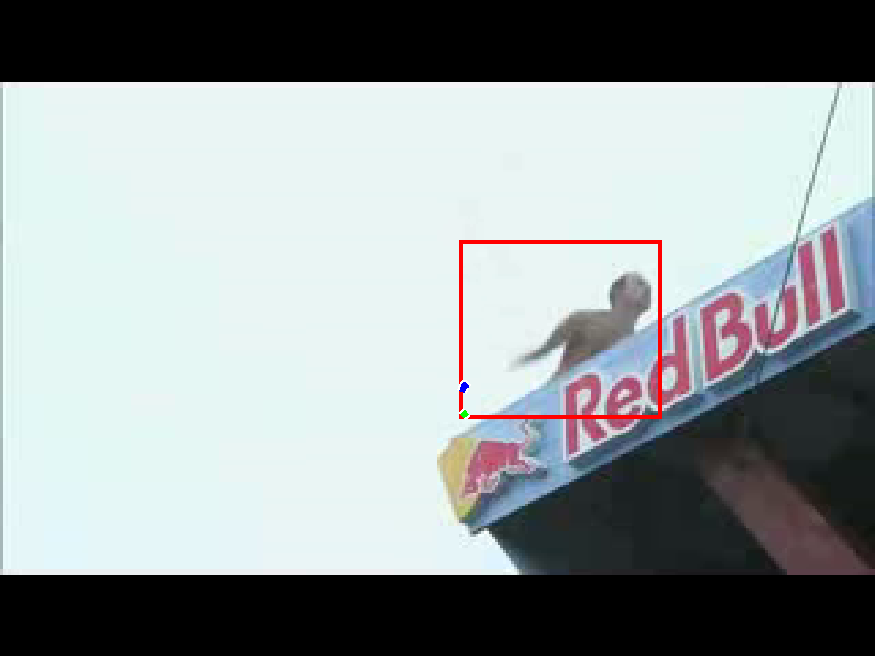}
			& \includegraphics[width=.18\textwidth, trim={0 42pt 0 42pt},clip] {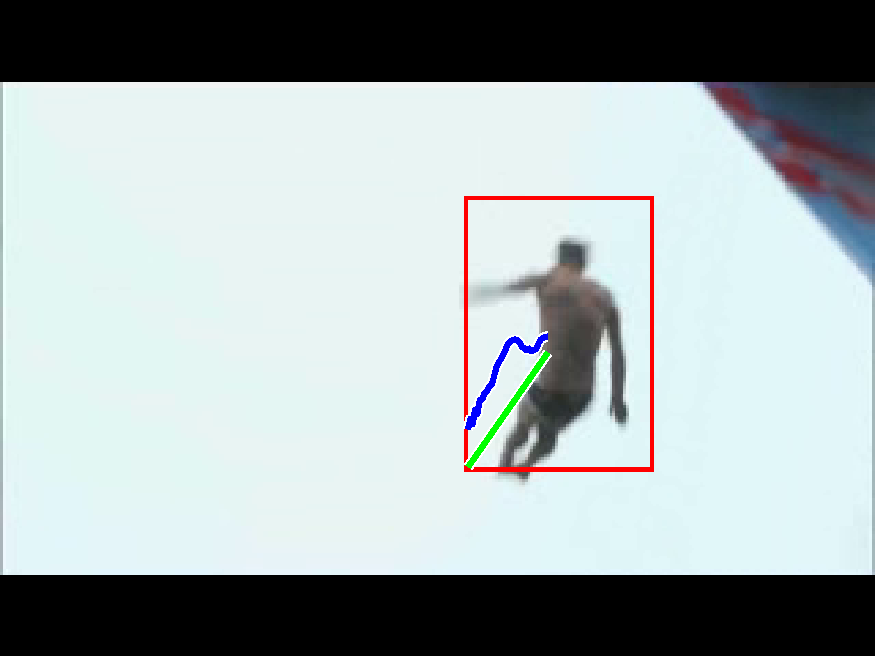}
			& \includegraphics[width=.18\textwidth, trim={0 42pt 0 42pt},clip] {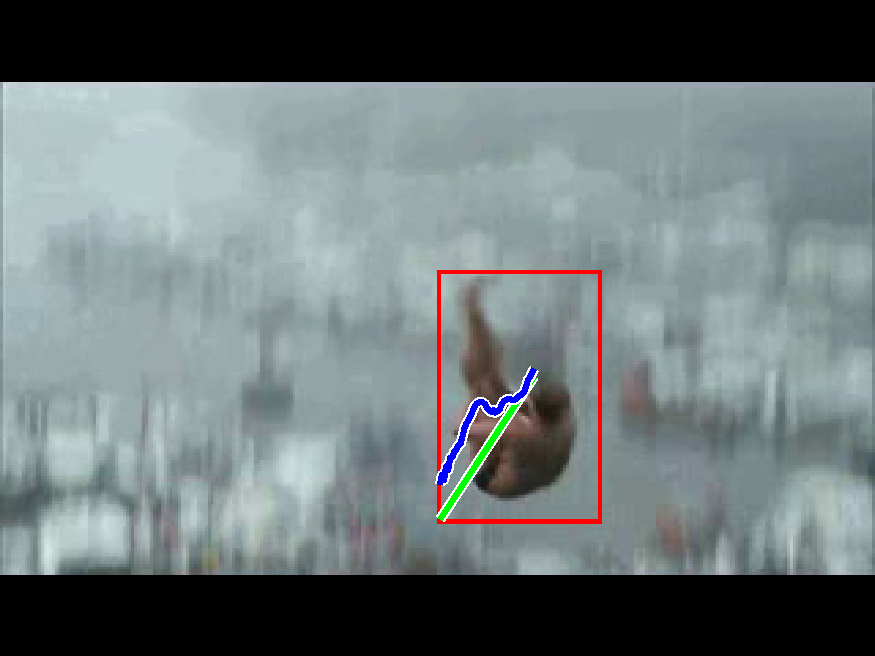}
			& \includegraphics[width=.18\textwidth, trim={0 42pt 0 42pt},clip] {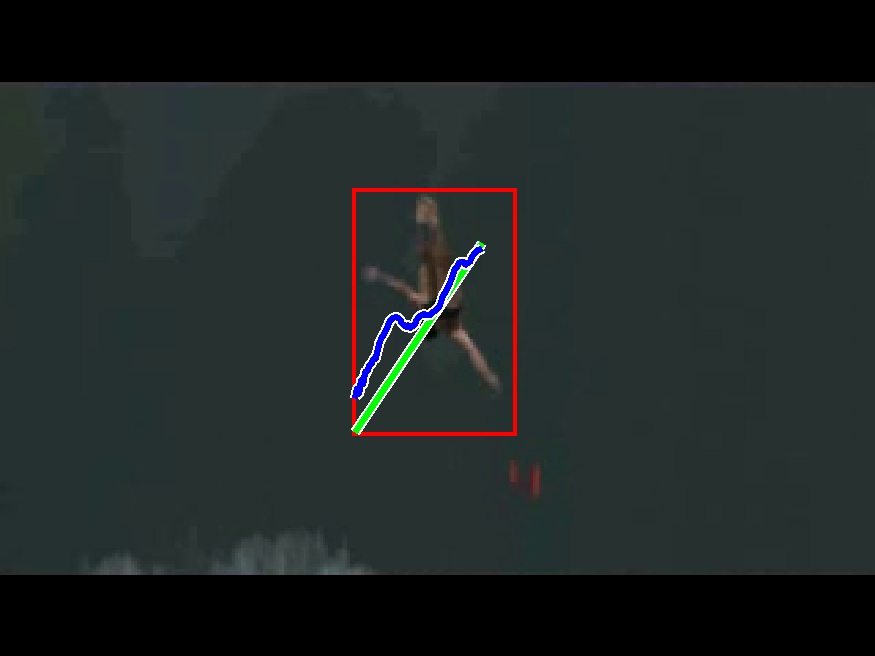}
			& \includegraphics[width=.18\textwidth, trim={0 42pt 0 42pt},clip] {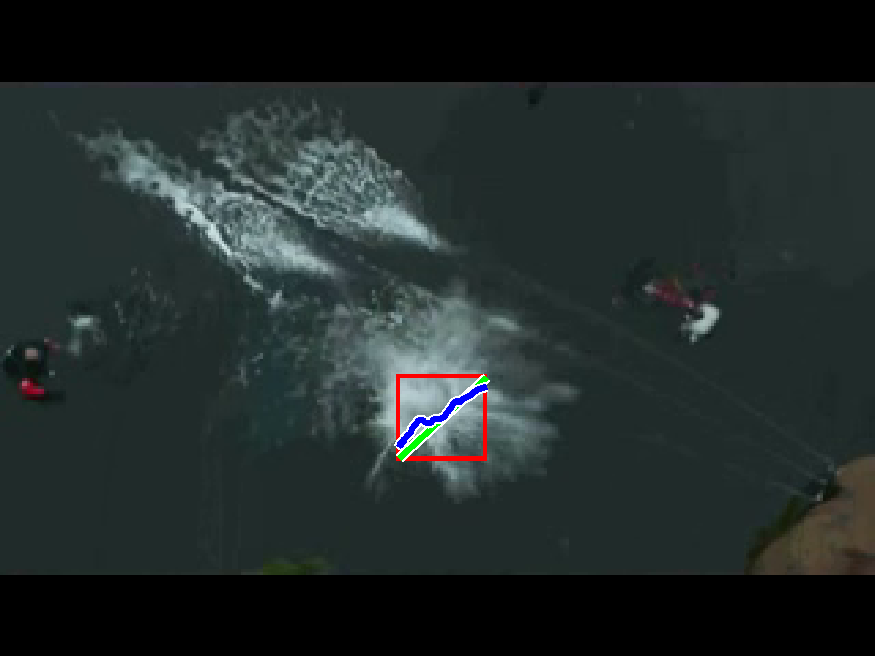}
		\end{tabular}
		
		\begin{tabular}{ccccc}
			\includegraphics[width=.18\textwidth, trim={0 22pt 0 62pt},clip] {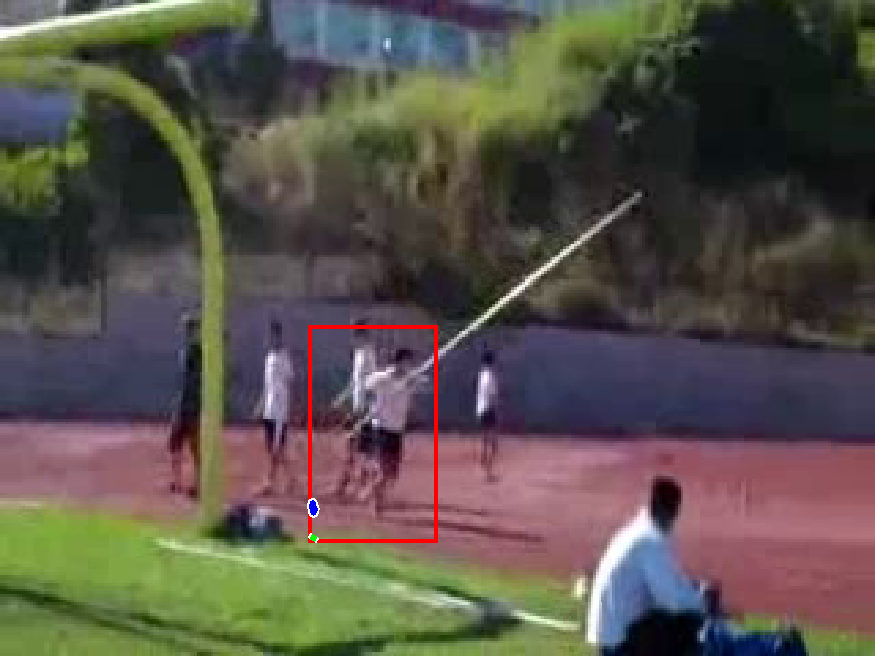}
			& \includegraphics[width=.18\textwidth, trim={0 22pt 0 62pt},clip] {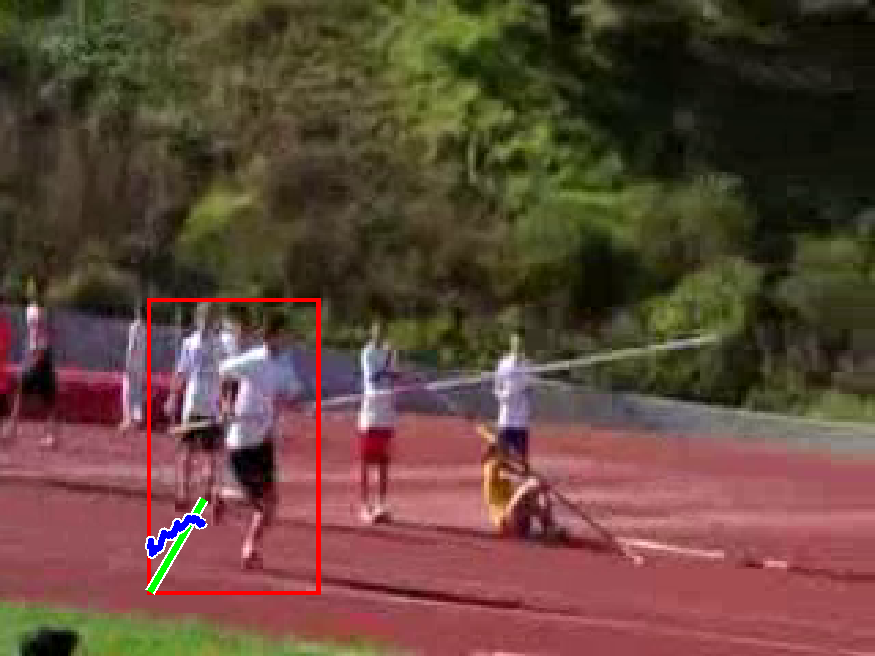}
			& \includegraphics[width=.18\textwidth, trim={0 22pt 0 62pt},clip] {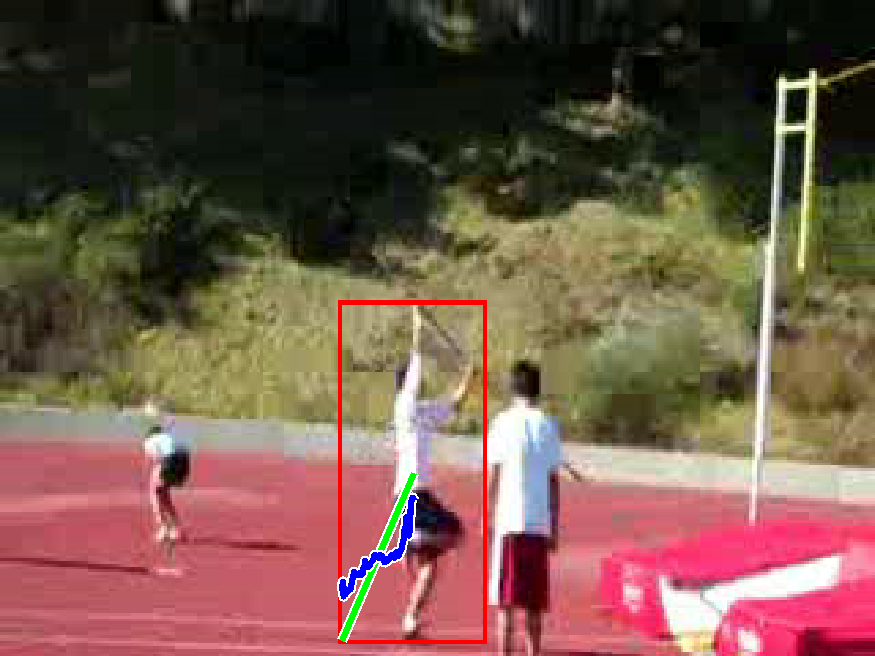}
			& \includegraphics[width=.18\textwidth, trim={0 22pt 0 62pt},clip] {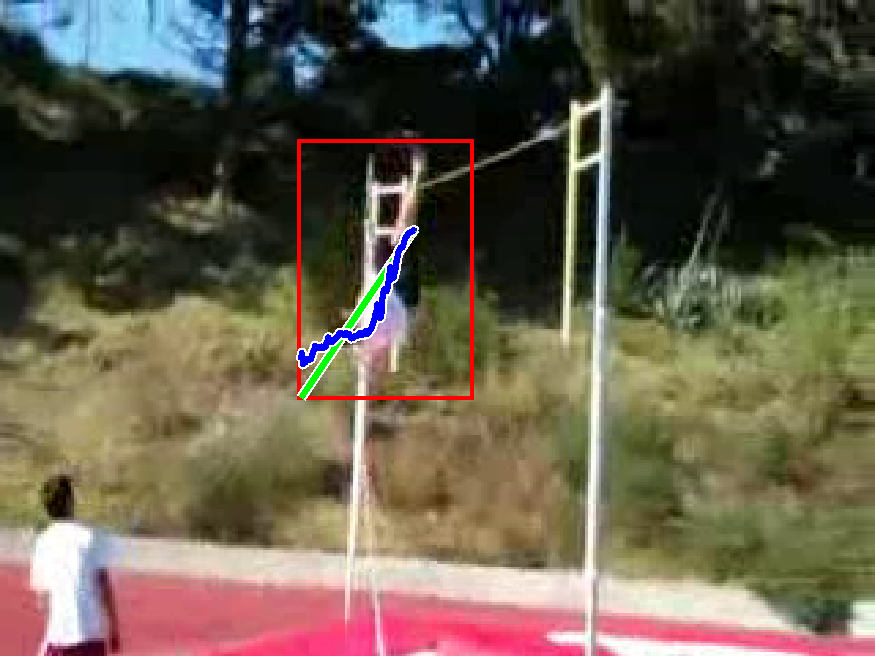}
			& \includegraphics[width=.18\textwidth, trim={0 22pt 0 62pt},clip] {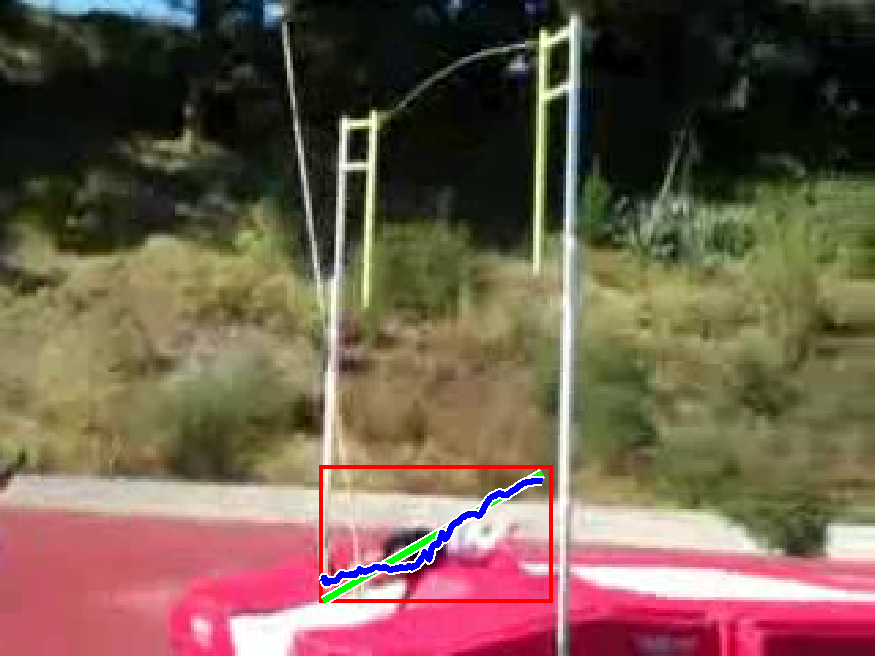}
		\end{tabular}
		
	\end{tcolorbox}
	%
	
	\begin{tcolorbox}[colframe=blue, left=-4pt, right=0pt,top=0pt, bottom=-4pt]
		\centering
		\begin{tabular}{ccccc}
			\includegraphics[width=.18\textwidth, trim={0 32pt 0 52pt},clip] {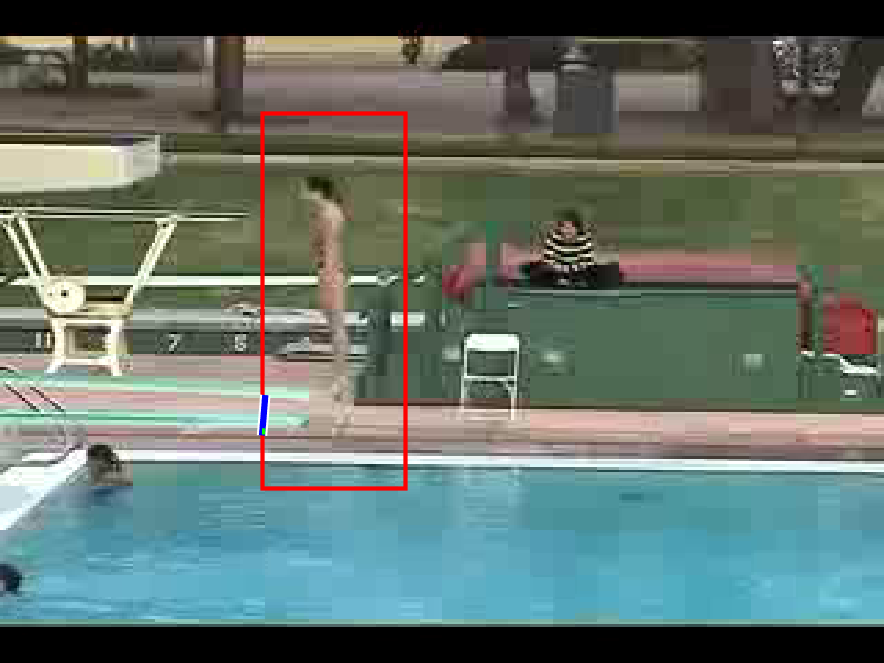}
			& \includegraphics[width=.18\textwidth, trim={0 32pt 0 52pt},clip] {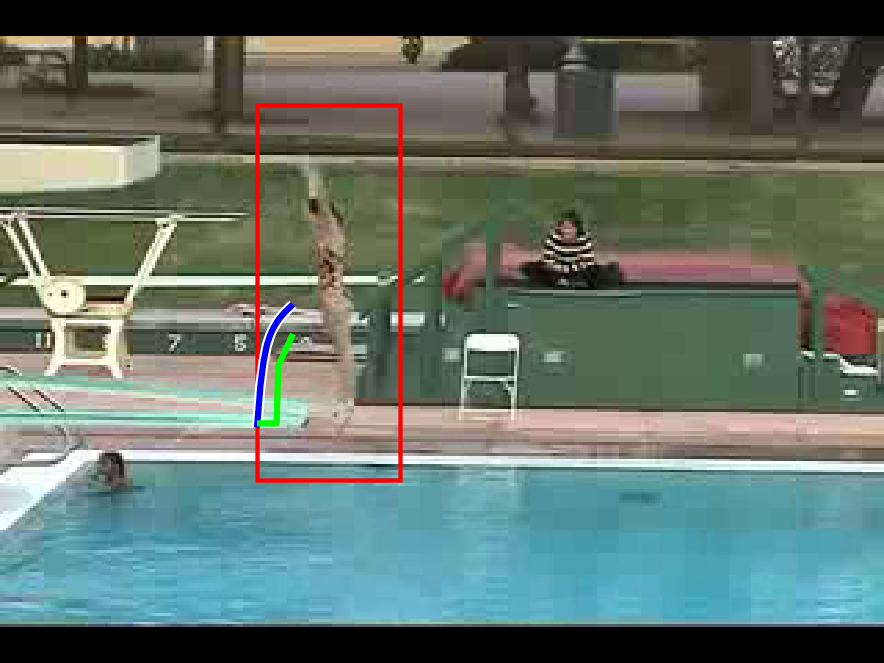}
			& \includegraphics[width=.18\textwidth, trim={0 32pt 0 52pt},clip] {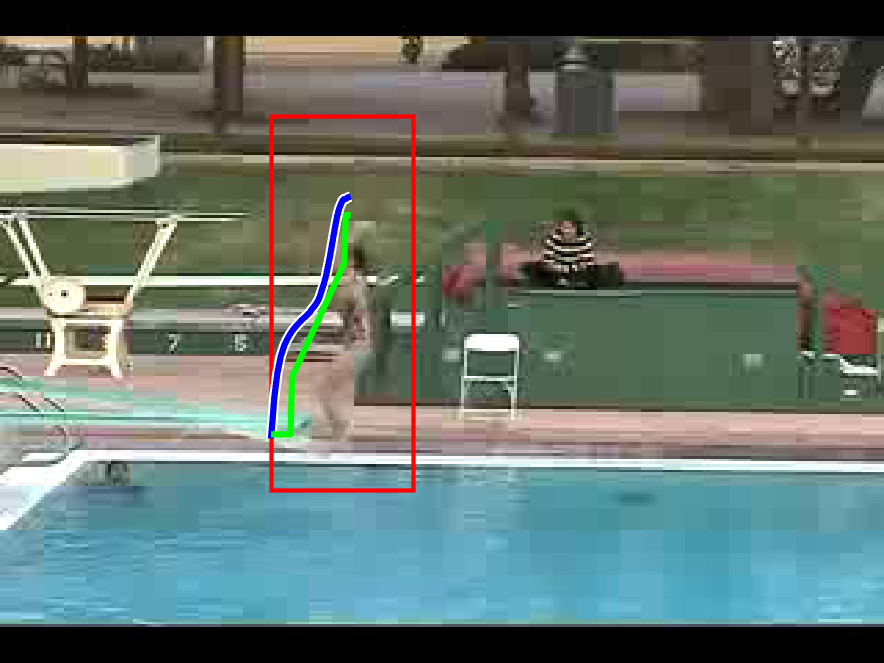}
			& \includegraphics[width=.18\textwidth, trim={0 32pt 0 52pt},clip] {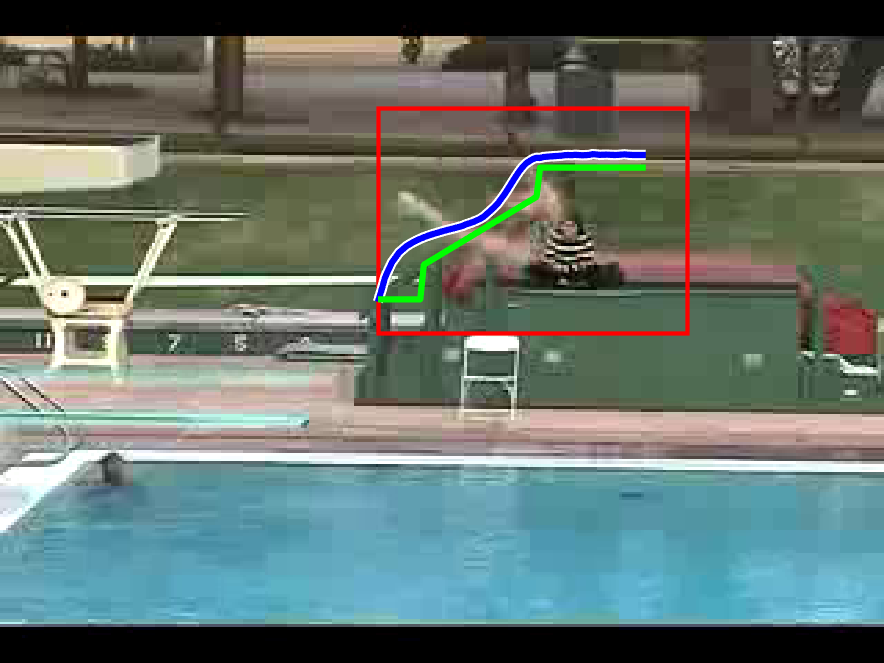}
			& \includegraphics[width=.18\textwidth, trim={0 32pt 0 52pt},clip] {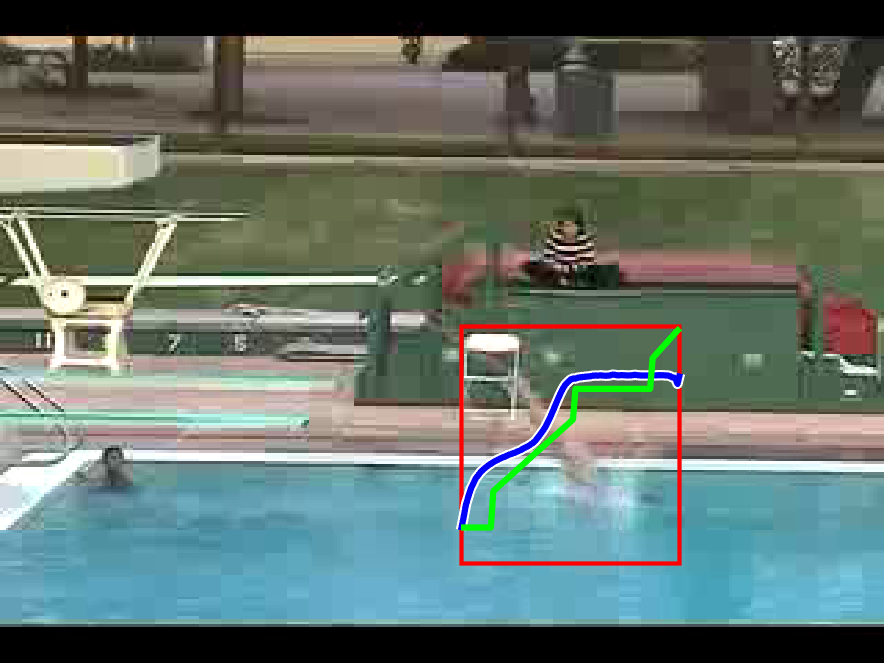}
		\end{tabular}
		
		\begin{tabular}{ccccc}
			\includegraphics[width=.18\textwidth, trim={0pt 72pt 80pt 62pt},clip] {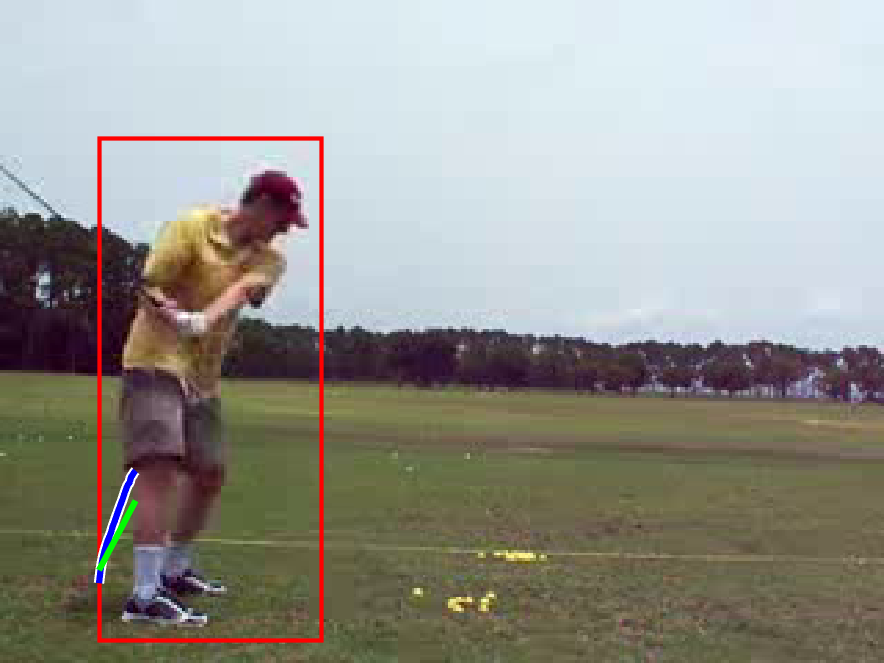}
			& \includegraphics[width=.18\textwidth, trim={0pt 72pt 80pt 62pt},clip] {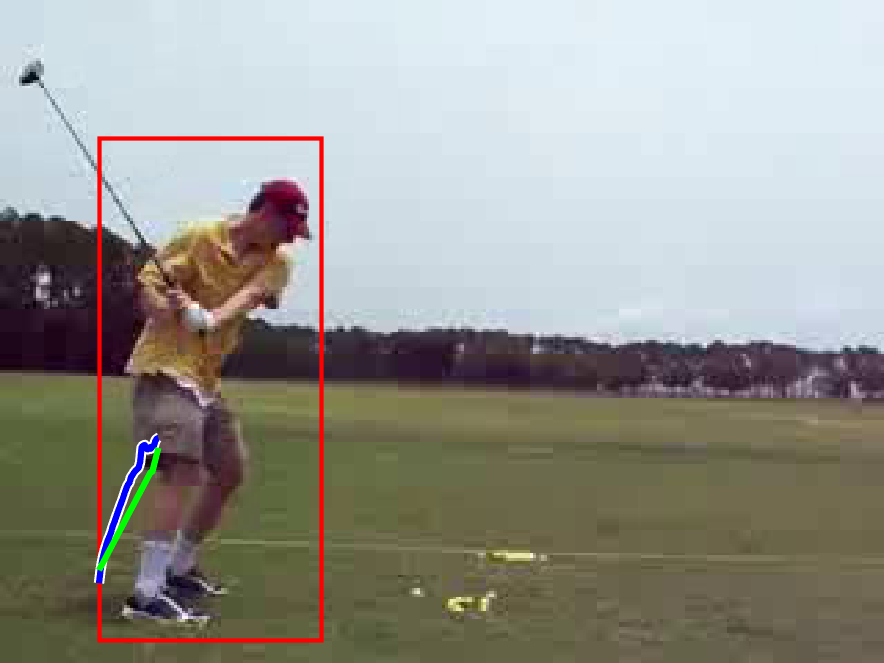}
			& \includegraphics[width=.18\textwidth, trim={0pt 72pt 80pt 62pt},clip] {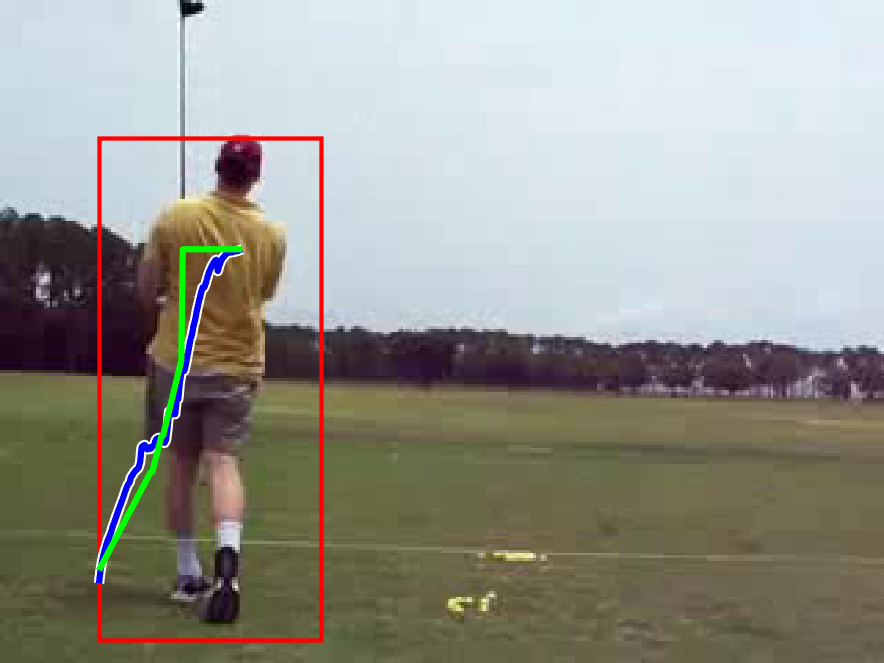}
			& \includegraphics[width=.18\textwidth, trim={0pt 72pt 80pt 62pt},clip] {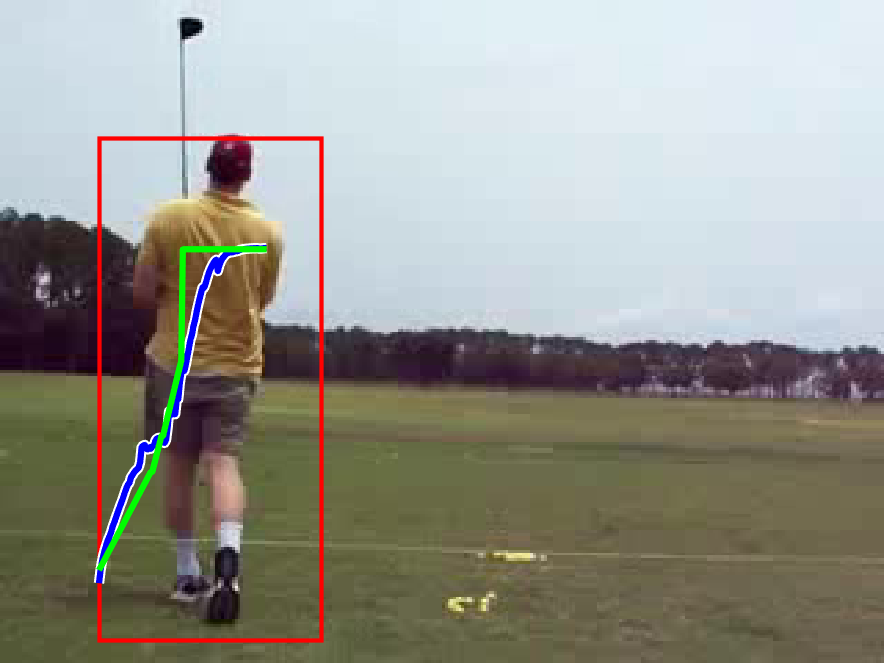}
			& \includegraphics[width=.18\textwidth, trim={0pt 72pt 80pt 62pt},clip] {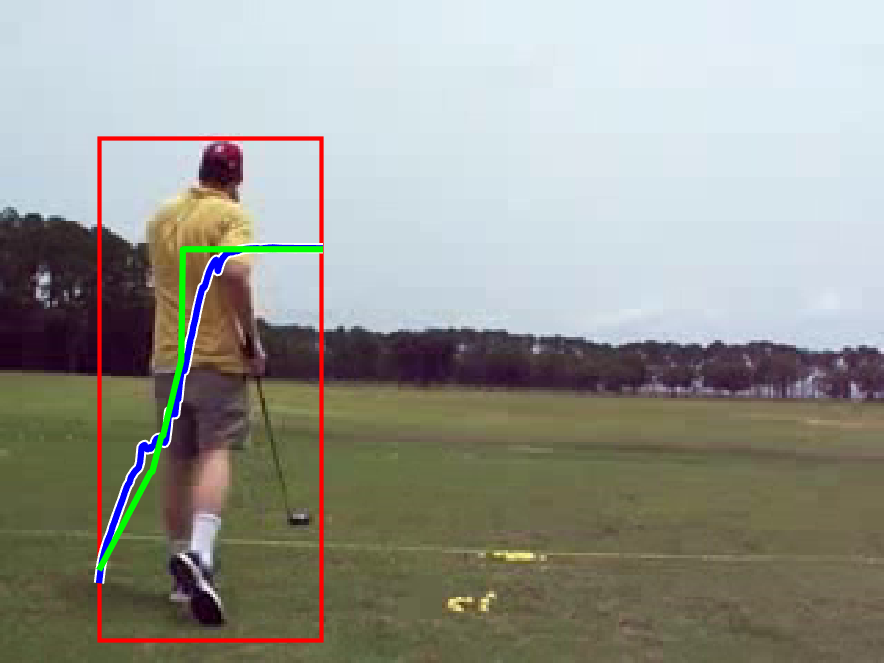}
		\end{tabular}
		
		\begin{tabular}{ccccc}
			\includegraphics[width=.18\textwidth, trim={55pt 72pt 25pt 62pt},clip] {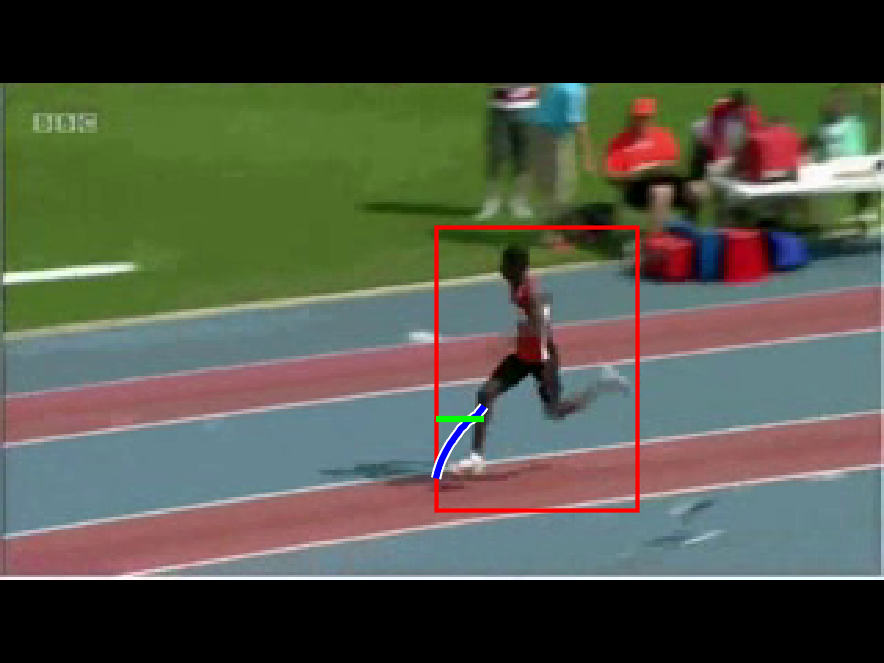}
			& \includegraphics[width=.18\textwidth, trim={55pt 72pt 25pt 62pt},clip] {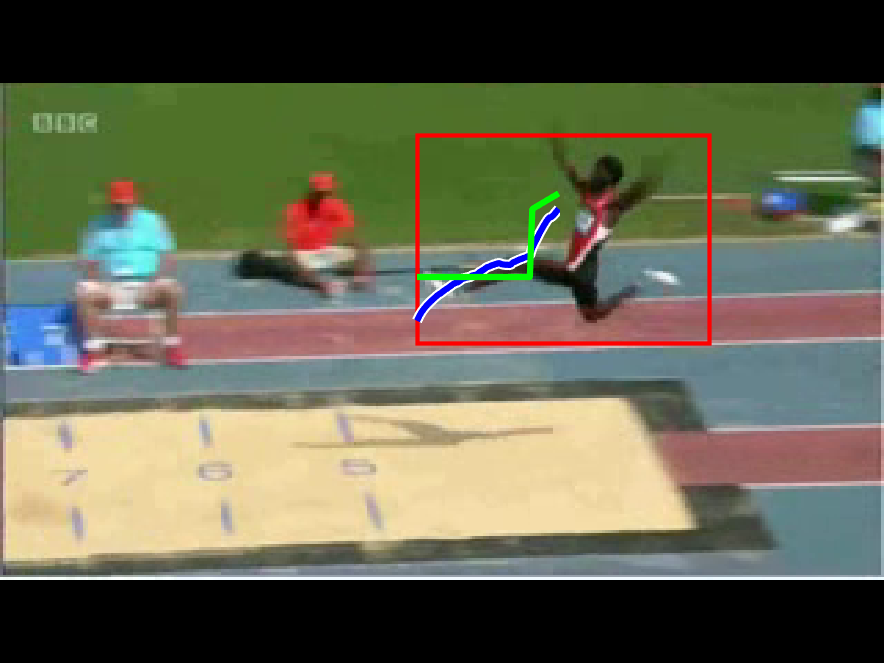}
			& \includegraphics[width=.18\textwidth, trim={55pt 72pt 25pt 62pt},clip] {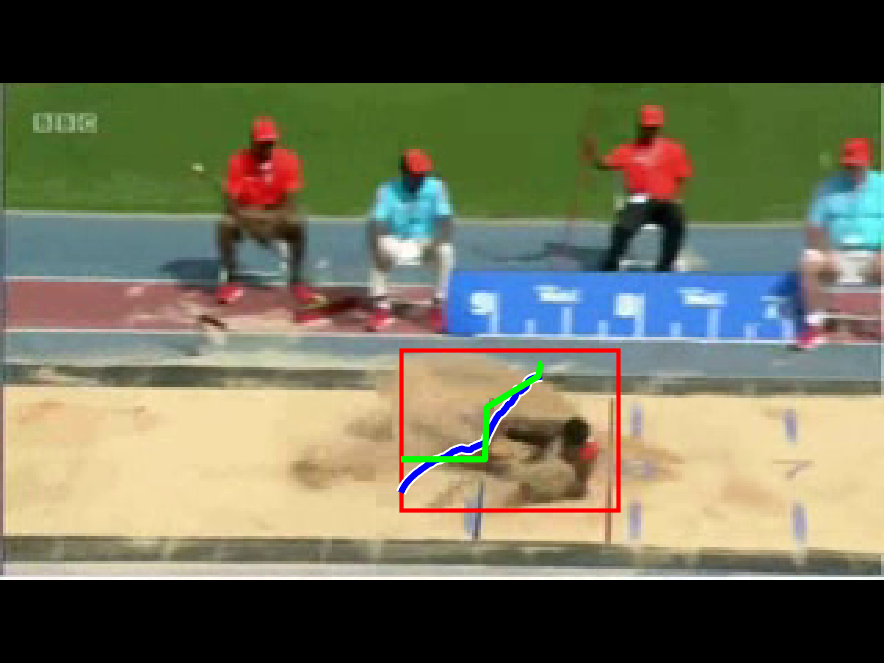}
			& \includegraphics[width=.18\textwidth, trim={55pt 72pt 25pt 62pt},clip] {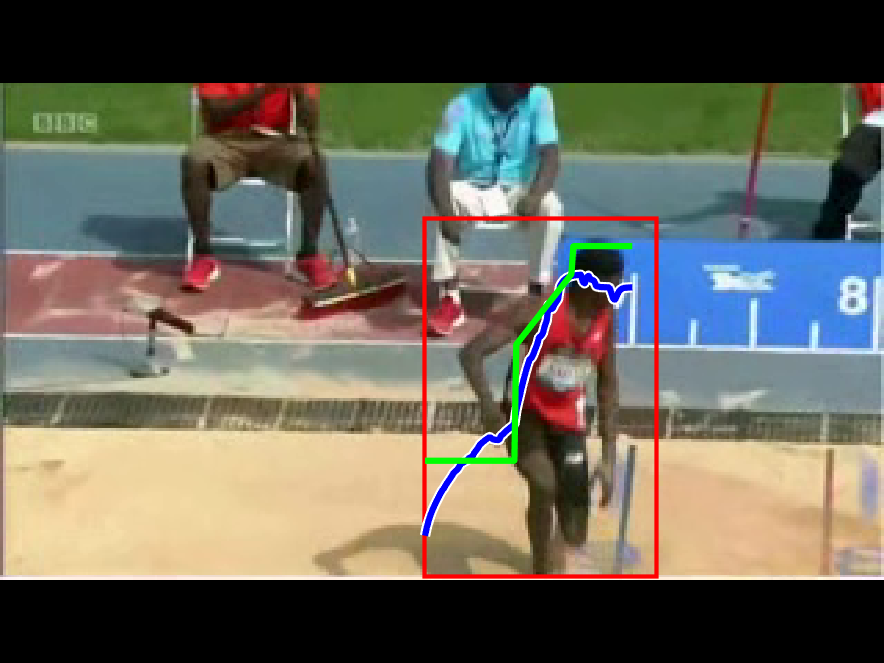}
			& \includegraphics[width=.18\textwidth, trim={55pt 72pt 25pt 62pt},clip] {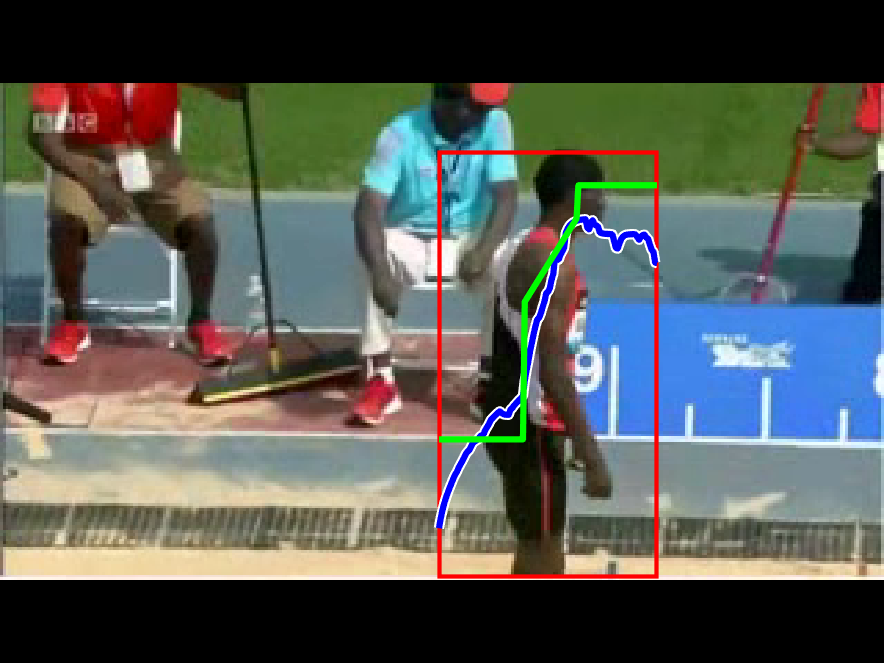}
		\end{tabular}
	\end{tcolorbox}
	\egroup
	
	\caption{Qualitative results with the linear based model (green frame) and phase-based model (blue frame). Each row represents the progression of an action. Progress values are plotted inside the detection box with time on the x axis and progress values on the y axis. Progress targets are shown in green and predicted progresses in blue.}
	\label{fig:crops_completion}
\end{figure*}

We inspect the results of ProgressNet trained with linear annotations and with phase-based annotations. Fig.~\ref{fig:crops_completion} shows some qualitative results with the two models on the UCF-101 dataset. It is interesting to notice how in some of the examples with the linear model, the predicted progress does not have a decise linear trend. Instead, it appears to follow the visual appearance of the action, changing its trend when there is a change in the semantics of the action. It appears that the model trained with the linear model has discovered to some extent the states that are made explicit with the phase-based annotations. For instance, the running phases in the first (\textit{LongJump}) and third row (\textit{PoleVaults}) clearly exhibit a different trend compared to the final parts of the actions, where the predictions increase more steadily towards completion.
This behavior becomes evident when action phases are taken into account by the model. It can be seen that in the three examples the predicted progress curves change their slope when entering into a different phase.


\section{Conclusion}\label{sec:conclusion}
In this paper we introduced ProgressNet, a model that can predict spatio-temporal localization of actions and at the same time understand their evolution by predicting progress online. We proposed two interpretations on progress: first, a linear one which has the advantage of being simple and applicable to any action detection dataset without any manual annotation; second, a phase-based interpretation which is more complex and requires a detailed manual annotation but at the same time provides a much richer and precise description of the ongoing action. To offer an appropriate study of action progress, we grounded our findings in the linguistics literature and terminology to characterize actions. In addition to our model, we proposed a Boundary Observant loss which helps to avoid trivial solutions by taking into account punctual phases that are present in the execution of the action. Experiments on two datasets showed that the proposed model is able to obtain promising performance.

\begin{acks}
The authors gratefully acknowledge the support of NVIDIA Corporation with the donation of the Titan XP GPU used for this research. Lamberto Ballan is partially supported by the PRIN 2017 project "PREVUE - PRediction of activities and Events by Vision in an Urban Environment".
\end{acks}

\bibliographystyle{ACM-Reference-Format}
\bibliography{progressnet_tomm}

\end{document}